\pdfoutput=1
\documentclass[11pt, dvipsnames]{article}

\usepackage[preprint]{acl}

\usepackage{times}
\usepackage{latexsym}
\usepackage{placeins}
\usepackage{mdframed}

\usepackage[T1]{fontenc}

\usepackage[utf8]{inputenc}

\usepackage{microtype}

\usepackage{inconsolata}

\usepackage[frozencache,cachedir=.]{minted}
\setminted{style=default}
\usepackage{listings}

\usepackage{graphicx}
\usepackage{array}
\usepackage{subcaption}
\usepackage{multirow}
\usepackage{comment}
\usepackage{booktabs}
\usepackage{multirow}
\usepackage{amsmath}
\usepackage{wrapfig}
\usepackage{breqn}
\usepackage{caption} 
\usepackage[most]{tcolorbox}
\usepackage{mdframed}
\usepackage{pifont}
\newcommand{\cmark}{\ding{51}}  
\newcommand{\xmark}{\ding{55}}  
\usepackage{fontawesome5}

\newenvironment{enu}{                   
     \parskip 0cm \begin{list}{}{\parsep 0cm \itemsep 0cm \topsep 0cm}}{
       \end{list}} 
       
\newcommand{\coloredcircle}[1]{%
    \textcolor{#1}{\faCircle}
}

\title{HypER: Literature-grounded Hypothesis Generation and Distillation with Provenance 
} 

\author{
  \textbf{Rosni Vasu\textsuperscript{1}},
  \textbf{Chandrayee Basu\textsuperscript{2}},
  \textbf{Bhavana Dalvi Mishra\textsuperscript{3}},
\\
  \textbf{Cristina Sarasua\textsuperscript{1}},
  \textbf{Peter Clark\textsuperscript{3}},
  \textbf{Abraham Bernstein\textsuperscript{1}}
\\
\\
  \textsuperscript{1}University of Zurich,
  \textsuperscript{2}Cornell University,
  \textsuperscript{3}Allen Institute for AI
\\
 \small{
   \texttt{rosni@ifi.uzh.ch}
 }
}

\begin{document}
\maketitle
\begin{abstract}
Large Language models have demonstrated promising performance in research ideation across scientific domains. Hypothesis development, the process of generating a highly specific declarative statement connecting a research idea with empirical validation, has received relatively less attention. Existing approaches trivially deploy retrieval augmentation and focus only on the quality of the final output ignoring the underlying reasoning process behind ideation. We present $\texttt{HypER}$ (\textbf{Hyp}othesis Generation with \textbf{E}xplanation and \textbf{R}easoning), a small language model (SLM) trained for literature-guided reasoning and evidence-based hypothesis generation. $\texttt{HypER}$ is trained in a multi-task setting to discriminate between valid and invalid scientific reasoning chains in presence of controlled distractions. We find that $\texttt{HypER}$ outperformes the base model, distinguishing valid from invalid reasoning chains (+22\% average absolute F1), generates better evidence-grounded hypotheses (0.327 vs. 0.305 base model) with high feasibility and impact as judged by human experts ($>$3.5 on 5-point Likert scale).

\end{abstract}

\section{Introduction}

\begin{figure*}
    \centering
    \includegraphics[scale=0.7]{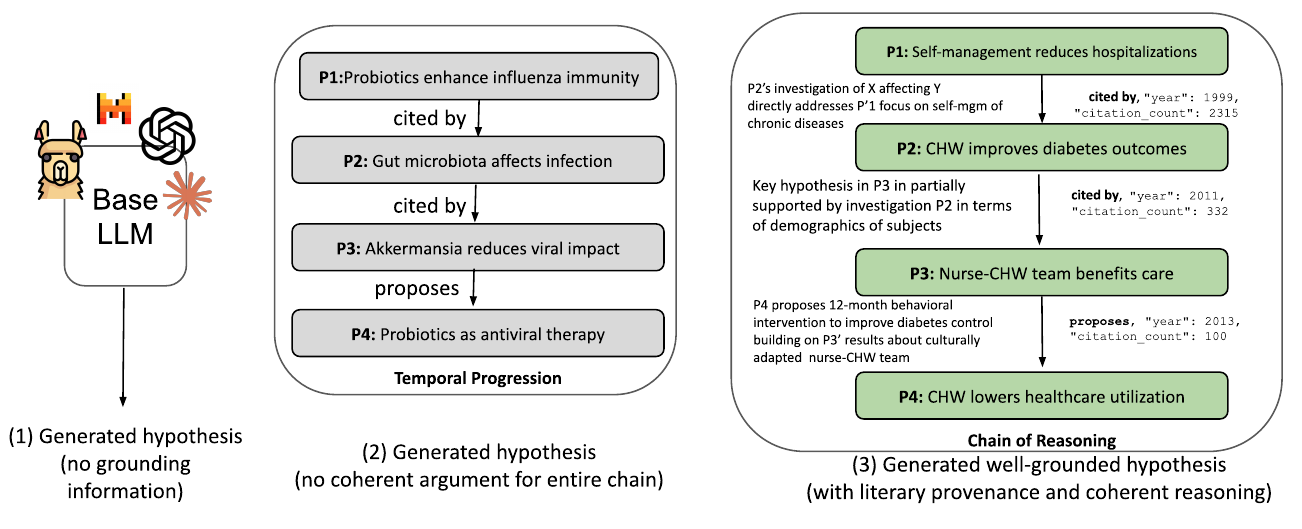}
    \caption{Comparison of hypothesis generation approaches: \textbf{(1) Base LLM} -- Generates ungrounded hypotheses. \textbf{(2) Citation Chain without Validation} -- Uses related papers but lacks a coherent argument across the chain. \textbf{(3) HypER (Ours)} -- Ensures hypotheses are well-supported with literature and logical progression. Please refer to Table~\ref{tab:dependency_chain} in Appendix for details of reasoning chains. 
}
    \label{fig:CoR}
\end{figure*}

Large Language Models (LLMs) have shown remarkable potential as AI scientists and research assistants, excelling in tasks such as knowledge acquisition from the scientific literature, idea generation, hypothesis development, experiment design, and data-driven verification~\cite{pu2024ideasynth, li2024chain, si2024can, wang2023scimon, lu2024ai, doi:10.1126/science.1165620, qi2024large}. One such scientific task is literature-based discovery (LBD), which aims to generate novel hypotheses by exploring connections within a large body of scientific literature. Techniques in LBD include structured causality investigations, including association rules, graph theoretics, and explicitly curated semantic relationships between concepts \cite{swanson1986undiscovered, xun2017generating}. In this work, we focus on LBD in the medical domain. 

In the medical domain, where evidence-based reasoning is the norm~\cite{DBLP:journals/corr/abs-1904-09612, bichindaritz1998case}, researchers require a clear provenance of ideas before committing to costly hypothesis development and validation~\cite{jing2024data, karunarathna2024evolution}. 
While traditional LBD methods provide structured pathways for discovery~\citep{swanson1986undiscovered, thilakaratne2019systematic}, their reliance on co-occurrence patterns limits the ability to capture evolving research trajectories. In contrast, LLMs enable the generation of creative, open-ended ideas by synthesizing diverse information~\citep{wang2023scimon}. However, this flexibility often comes at the cost of interpretability and grounding in scientific evidence, two attributes essential for real-world use in clinical and biomedical research.

Existing LLM-based approaches to scientific hypothesis generation, such as ResearchAgent~\citep{baek2024researchagent}, Acceleron~\cite{nigam2024interactive}, SciMuse~\cite{gu2024generation}, and SciMON~\citep{wang2023learning} treat the task as conditional generation over retrieved literature. Unlike traditional LBD systems, these models lack a structured approach to literature organization. A common practice in literature review is to organize prior work chronologically to discover trends, uncover key milestones, and build knowledge. Recent work demonstrated the effectiveness of this approach in AI assisted idea generation, e.g., inspirations presented as chains of ideas or paths connecting concepts in a Knowledge Graph (KG) was reported to improve the quality of research ideas. \cite{li2024chain, ghafarollahi2024sciagents}. 

We build on this idea of structured representations to bridge the gap between traditional LBD and LLM-based hypothesis generation. However, rather than imposing structure only at inference time, we argue that scientific AI assistants should be trained to organize and reason over the literature, mimicking real-world scientific inquiry. In this paper, we ask: \textit{How can we train an LLM to navigate the noisy literature and generate novel and impactful ideas that are grounded in a solid understanding of existing work.}. 

To address this question, we develop \texttt{HypER}, a small language model (SLM) trained for \emph{literature-guided reasoning} and \emph{evidence-based hypothesis generation}, focusing on fine-grained logical connections between arguments in scientific abstracts rather than collating ideas by surface-level similarity commonly used in recent work. We first validate a teacher LLM's capability to extract these dependencies. Then using the validated teacher, we contribute a novel dataset of temporal chains (sequences of article abstracts) where each node is \emph{inspired by} or \emph{dependent on} its predecessor, reflecting the evidence-driven nature of scientific discovery. To account for the real-world challenges, we simulate varying levels of noise in the literature with carefully curated controlled distraction articles. While building these chains requires costly citation graph traversal and numerous LLM calls, we distill this process into an SLM fine-tuned via multitask learning. \texttt{HypER} is trained to discriminate between valid and invalid chains and to integrate this reasoning with the ideation of evidence-based hypotheses. This paper makes the following contributions:
 
\begin{enu}
\item \textbf{Task:} We formalize a new literature-grounded scientific hypothesis generation task that goes beyond surface-level similarity-based linking by explicitly validating reasoning chains using relevance scoring to ensure logical coherence and progression across cited papers.
\item \textbf{Framework:} We propose a multitask framework that explicitly supervises the scientific reasoning process via two classification tasks: one-hop paper-paper relevance and validity of multihop chains. Hypothesis generation is performed at inference time, conditioned on validated reasoning chains.
\item \textbf{Dataset:} We construct a dataset of 3,523 reasoning chains derived from 359 core valid chains, with fine-grained relevance labels and curated invalid samples, using LLM-based scoring validated through expert evaluation.
\item \textbf{Model:} We fine-tune small instruction-tuned LLMs (e.g., Phi-3-mini-128k-instruct), \texttt{HypER}, achieving strong performance across all tasks. 
\item \textbf{Evaluation: } We present a comprehensive evaluation of base vs \texttt{HypER}. For classification, we report accuracy and F1; for generation, we assess novelty, clarity, and groundedness via automated and expert review. \texttt{HypER} outperformed the base model at distinguishing valid vs. invalid reasoning chains (+22\% average absolute F1) and generates more evidence-grounded hypotheses (0.327 vs. 0.305 base model) with high feasibility and impact as judged by human experts including clinicians and biomedical researchers ($>$3.5 on 5-point Likert scale), in some cases anticipatory of recent studies (Section~\ref{result:human} and Appendix~\ref{appendix: expert_analysis}.)
\end{enu}

Although we focus on the medical domain for its strong emphasis on evidence-based reasoning, we believe the framework has potential applicability to other scientific fields. An important avenue for future work is to validate this generalizability through experiments in additional domains.

\section{HypER}\label{sec:problem}

\begin{figure*}[t]
  \centering
  \begin{subfigure}[t]{0.70\textwidth}
    \centering
    \includegraphics[width=1\linewidth]{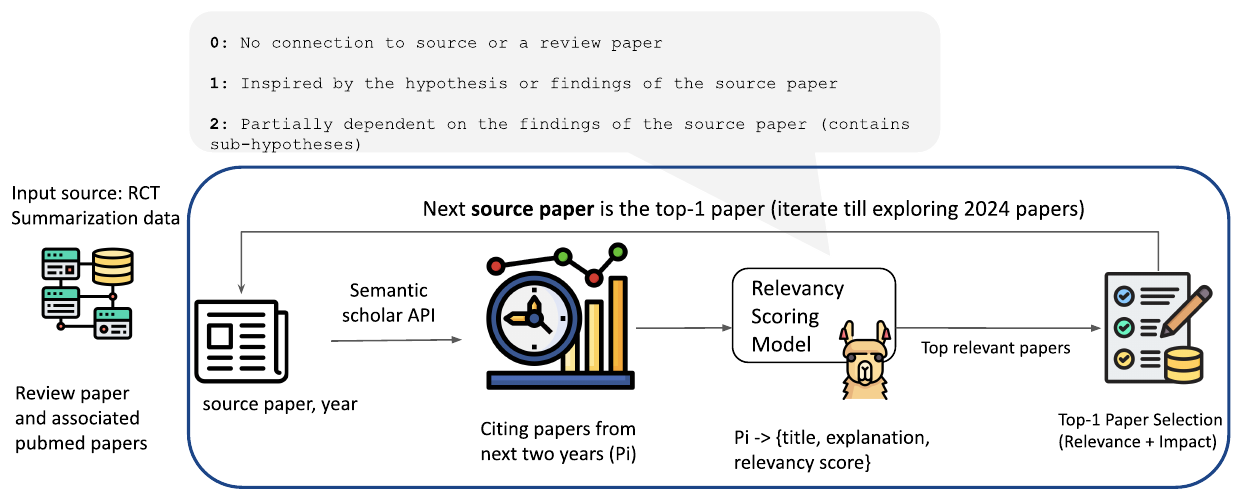}
    \caption{}\label{fig:pipeline}
  \end{subfigure}\hfill
  \begin{subfigure}[t]{0.25\textwidth}
    \centering
    \includegraphics[width=0.8\linewidth]{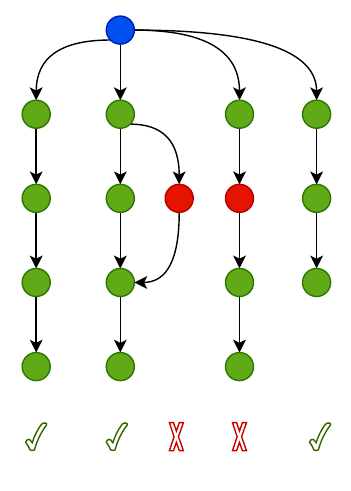}
   \caption{}
    \label{fig:chain-graph}
  \end{subfigure}
  \caption{(a) Pipeline for constructing reasoning chains from RCT summarization data. The process iteratively retrieves citing papers, evaluates their relevance, and constructs a literature path of evidence-supported hypothesis reasoning. (b) An example sub-graph rooted from source paper $p_k$ (\coloredcircle{blue}) and multiple valid and invalid chains associated with it within it. All the \coloredcircle{green} are valid papers citing the previous papers and \coloredcircle{red} are the papers which are irrelevant to the previous paper but are cited them and might be sharing terminologies. }
  \label{fig:combined}
\end{figure*}

\subsection{Problem formulation}\label{sec:problem}
Our goal is to ensure that the reasoning paths not only support the generated hypothesis but also provide a clear and scientifically sound rationale, mimicking the thought processes of expert scientists.

We define the \emph{scientific literature graph} as $\mathcal{CG} = (\mathcal{P}, \mathcal{E})$, where $\mathcal{P}$  is the set of papers (nodes) and $\mathcal{E}$ the set of citation edges. An edge $(p_i, p_j) \in \mathcal{E}$ indicates that paper $p_i$ cites $p_j$. Each paper $ p_i \in \mathcal{P} $ is associated with a key hypothesis $ h_i$. The \emph{temporal reasoning chain} denoted as $\mathcal{C} = \{p_1, p_2, \dots, p_n\}$, is a sequence of papers, where $p_1$ is the source (anchor) paper, $\{p_2, \dots, p_{n-1}\}$  represent papers that appear in chronological order, reflecting the progression of scientific discovery. Each paper $p_i 
\in \mathcal{C}$ cites the previous paper $p_{i-1}$ and establishes a logical connection. $p_n$ represents a target paper (e.g., with target hypotheses). Unlike a raw citation graph, the reasoning chains we construct form a structured subset $\mathcal{VCG}$ of $\mathcal{CG}$, where edges are validated based on citation links and scientific dependencies (i.e., \emph{inspired by} or \emph{dependent on} the findings) rather than solely citation links---which could indicate a broader relationship (e.g., a cited paper to support a claim, a cited paper that works on the same area but has unrelated hypotheses). This ensures that $\mathcal{C}$ captures meaningful reasoning paths rather than arbitrary citation relationships. See Table~\ref{tab:dependency_chain} and~\ref{tab:dependency_chain2} in Appendix for details of reasoning chain examples.

We utilize our collection of reasoning chains $\mathcal{VCG}$ to train a small language model, called $\texttt{HypER}$, on three interconnected reasoning tasks that enhance its hypothesis generation capabilities. By fine-tuning a single model across multiple tasks, leveraging shared representations, we enable $\texttt{HypER}$ to conduct literature-based-discovery in a real world setting.

\subsection{Multi-task objectives}\label{sec:multi-task}

$\texttt{HypER}$ is fine-tuned on the following tasks:

\paragraph{One-hop relevance classification ($\texttt{1-hop}$):} given a \textit{source paper} and a \textit{target paper}, the model predicts a \textit{relevancy score}, similar to the one described in our data generation pipeline (details in Section~\ref{chain_construction}). Here, the model assigns the relevancy score to the target paper based on its scientific dependence on a source paper (scored as 0: irrelevant, 1: inspired, or 2: dependent), focusing on fine-grained, local dependencies in the literature graph. 
\paragraph{Multi-hop agnostic chain validation ($\texttt{multi-hop-A}$):} 
Given a \textit{reasoning chain} (a sequence of temporally ordered papers), the model determines whether the chain is \textit{valid} or \textit{invalid}. If invalid, it identifies the specific breakpoints (paper nodes in the chain) where inconsistencies occur in the logical progression. This task improves the model's ability to differentiate valid from noisy reasoning paths. 
We argue that a model should be able to identify inconsistencies in scientific reasoning irrespective of the target hypothesis based on the coherence of the argument. 
\paragraph{Multi-hop contextual chain validation ($\texttt{multi-hop-C}$):} Given a \textit{reasoning chain} (a sequence of temporally ordered papers), and target hypotheses that leads to, the model determines whether the chain is {invalid}. If invalid, it identifies the specific breakpoints (paper nodes in the chain) where logical inconsistencies occur. This ensures that hypothesis generation by the $\texttt{HypER}$ is supported by well-structured and scientifically sound reasoning.

This multi-task setup enables joint training with task-specific instruction tuning and a shared model backbone to improve generalization across reasoning levels. The one-hop relevance scoring task helps the model capture fine-grained scientific dependencies, which contributes to its ability to evaluate multi-hop reasoning chains. 

\subsection{Hypothesis Generation with $\texttt{HypER}$}
While \texttt{HypER} is trained to validate reasoning chains, its ultimate goal is to generate well-grounded scientific hypotheses. By distinguishing valid from invalid reasoning chains, the model identifies coherent reasoning chains within noisy literature graphs and uses them as scaffolds for hypothesis generation. To demonstrate the above, we prompted our fine-tuned model with a noisy literature subgraph. Given a noisy subgraph $\mathcal{G}_{\mathcal{C}_i}$, and a task-specific prompt $\mathcal{T}$, the model outputs the valid reasoning chain 
$\mathcal{C}_i$, a rationale or explanation $\mathcal{R}_i$ and the final hypothesis $h_{i}$.

\section{Constructing a Ground Truth for Literature-Guided Reasoning Chains}\label{chain_construction}
To facilitate hypothesis generation, grounded in the logically connected chain of scientific developments, we construct a structured subset of scientific literature graph $\mathcal{G} \subset \mathcal{CG}$, consisting of valid reasoning chains along with distraction chains (right of Figure~\ref{fig:combined}). These chains serve as structured representations of the logical steps connecting existing knowledge to new hypotheses. The pipeline for constructing valid reasoning chains is designed to iteratively retrieve, score, and validate scientific literature, ensuring that each reasoning path is logically grounded in evidence as shown in the left of Figure \ref{fig:combined}. Below, we describe each step of the pipeline in detail:

\begin{table*}[ht!]
\centering
\renewcommand{\arraystretch}{1.2}
\small
\begin{tabular}{lcccc}
\hline
\textbf{Type of Chain}        & \textbf{Disruption Level} & \textbf{Number of Chains} & \textbf{Mean Length (Min, Max)} & \textbf{Score 2 Fraction (Mean)} \\ \hline
\textbf{Valid Chains}         & N/A                       & 379                         & 9.04 (1, 27)                    & 0.71                          \\
                            \hline
\textbf{Easy Chains}          & 10\% Replacements         & 175                         & 13.88 (10, 27)                 & 0.65                            \\ 
                              & 20\% Replacements         & 342                         & 11.67 (5, 27)                  & 0.55                           \\ 
                              & 30\% Replacements         & 305                         & 12.12 (4, 27)                  & 0.49                           \\ 
                              & 40\% Replacements         & 295                         & 12.31 (3, 27)                  & 0.41                           \\ 
                              & 50\% Replacements         & 67                         & 16.01 (12, 27)                 & 0.40                            \\ \hline
                            & Total                       & 1184                         & 12.52 (3, 27)                  & 0.55                           \\ \hline
\textbf{Hard Chains}          
                            & N/A                      & 455                          & 9.97 (2, 28)                  & 0.62                            \\ \hline

\end{tabular}
\caption{Statistics of valid and invalid chains. Easy chains are invalid chains with varying disruptions.}
\label{tab:chains_stats}
\end{table*}

\paragraph{Step 1: Data Preparation} 
The process begins with sampling a set of papers from a dataset~\citep{AMIA-summarization-2021} of randomized controlled trial (RCT) summaries. We utilized the dataset by~\citet{AMIA-summarization-2021}, which is based on systematic reviews curated by experts. Each review is relevant to a clinical question and linked to multiple PubMed papers that serve as potential initial source papers. We select a source paper $ p_k$- either the latest or most cited and note its publication year. Sub-discipline selection details are provided in Appendix~\ref{appendix:inputRCT}.

\paragraph{Step 2: Citation Graph Retrieval} Using the Semantic Scholar API, we retrieve papers citing $p_k$ within a two-year window ($year \rightarrow year+2$), grouped into batches of 10 to fit within LLM context limits.

\paragraph{Step 3: Relevancy Scoring for a Paper}
Each paper is scored using a Llama-3.1-70B model (prompt in Appendix~\ref{appendix:prompt_llama_score}) with a relevance label: \texttt{0} (irrelevant), \texttt{1} (inspired), or \texttt{2} (dependent), based on its connection to the source paper’s hypothesis or findings. The model also outputs a brief explanation of its relevancy and the paper title.

\paragraph{Step 4: Top paper selection} For each paper chunk, the top 3 relevant papers are identified based on their relevancy score in the range [1, 2]. Papers with higher citation counts and relevancy scores of 2 (only considered score 1 otherwise) are prioritized. Only papers with valid scores (e.g., [1, 2]) are retained for further processing in the chain using a relevance impact score, that considers the relevancy (70\%) and impact (30\%) using the citation count. This approach ensures that highly relevant and impact papers are prioritized, mitigating coverage gaps that could disrupt the reasoning chain if lower-impact papers were included.

\paragraph{Step 5: Iterative Reasoning Chain Construction} The pipeline iteratively selects the top paper from the relevant papers. This paper becomes the new source paper $p_{k+1}$, and the process is repeated to retrieve its citing papers. The loop continues until a terminal condition is met, such as reaching the final target year (e.g., 2024).
\\
The reasoning chain is constructed as a sequence of papers {$p_k, p_{k+1}, \ldots, p_{kn}$}, where each node represents a relevant paper contributing to the idea.

\subsection{Generating causal chains of literature with teacher LLM}
A key step of our reasoning chain construction is evaluating whether a paper is inspired by, or depends on the findings of the previous paper in the chain. Unlike prior works like SciMon~\cite{wang2023scimon} and COI~\cite{li2024chain}, which rely on simple cosine similarity, we explicitly validate these fine-grained dependencies. The fidelity of silver data --- constructed ground truth --- relied on evaluating the hypothesis that LLMs can identify strong dependencies between scientific contributions. 

Our dataset consists of more relevant logical connections between papers along the chains. To establish this connection, we leverage large LLM's built-in reasoning capabilities. Specifically, our relevancy scoring depends on how well an LLM can discriminate between strong and weak one-hop connections between two chronologically ordered papers. \textbf{Manual Quality Assurance:} We validate the model's capability using self-consistency runs (see appendix~\ref{appendix:llamaScoring}) of 50 samples and comparison of the majority votes with human judgment. The expert annotated relevancy scores compared against majority vote LLM scores, achieved an average Cohen's Kappa of $0.429 \pm 0.065$ and a percentage agreement of $62.74 \pm 4.24\%$, indicating moderate agreement (details in Appendix~\ref{appendix:llamaScoring}). We constructed a total of \textit{379 reasoning chains}, each representing a structured progression of ideas connecting a source paper to a hypothesis. We present a summary statistics of our reasoning chains in the Appendix \ref{chain_stats}.

\subsection{Negative Sampling Strategies}
To create a dataset for the reasoning path validity task, we additionally need negative examples.  Invalid reasoning chains are generated through the following strategies (illustrated on the right of Figure~\ref{fig:combined} (b)): (1) \textbf{Swapping intermediate nodes (easy negative sampling):} Intermediate nodes in the reasoning chain are replaced with irrelevant or unrelated nodes. We selected the replacement nodes carefully from a pool of candidate papers with relevance $0$ from the same citing year to ensure the structural similarity is maintained while introducing invalid reasoning. For a valid chain, $\mathcal{VC} = {p_1 \rightarrow p_2 \rightarrow p_3 \rightarrow p_4}$ an invalid chain could be $\mathcal{VC}' = {p_1 \rightarrow p_5 \rightarrow p_3 \rightarrow p_4}$, where $p_5$ is irrelevant to $p_1$. This approach generates invalid chains with varying levels of noise by progressively replacing 10\% to 50\% of the intermediate nodes in the reasoning path. 
(2) \textbf{Random breaks in the chain (hard negative sampling):} In this strategy, we disrupt the reasoning chain by introducing random breaks (replacing 1 or 2 intermediate nodes with unrelated relevance $0$ nodes), resulting in disjoint subchains. After each break, the chain resumes as a valid chain, which preserves temporal and logical progression. This makes the invalid chain a hard negative with partial coherence and carefully crafted disruption. In this method, we introduce a level of randomness by removing the fixed target node assumption, unlike previous approaches where the reasoning chain is fixed between a source and target node. For instance, consider the following valid reasoning chain: $ \mathcal{VC} = \{p_1 (2001) \rightarrow p_2 (2004) \rightarrow p_3 (2007) \rightarrow p_4 (2011) $ $
\rightarrow p_5 (2015) \rightarrow p_6 (2020) \rightarrow p_7 (2024)\}$, we introduce a break after $p_3 (2007)$, we replace $p_4 (2011)$ with an irrelevant paper, $q_1$ followed by a coherent sub-chain of papers that are irrelevant to the initial valid chain $\mathcal{VC}' = \{p_1 (2001) \rightarrow p_2 (2004) \rightarrow p_3 (2007) \rightarrow q_1 (2011) $ $\rightarrow q_2 (2014) \rightarrow q_3 (2018) \rightarrow q_4 (2024)\}$
These strategies ensure diverse negative samples, enabling models to robustly differentiate between valid and invalid reasoning chains. We formally denote easy negative chains as \textit{invalid-easy} and hard negative chains as \textit{invalid-hard} for later reference.
\noindent
The distribution of valid, easy, and hard chains, along with their associated statistics, is summarized in Table~\ref{tab:chains_stats}.
\FloatBarrier

\begin{figure}[ht]
    \centering
    \includegraphics[width=\linewidth]{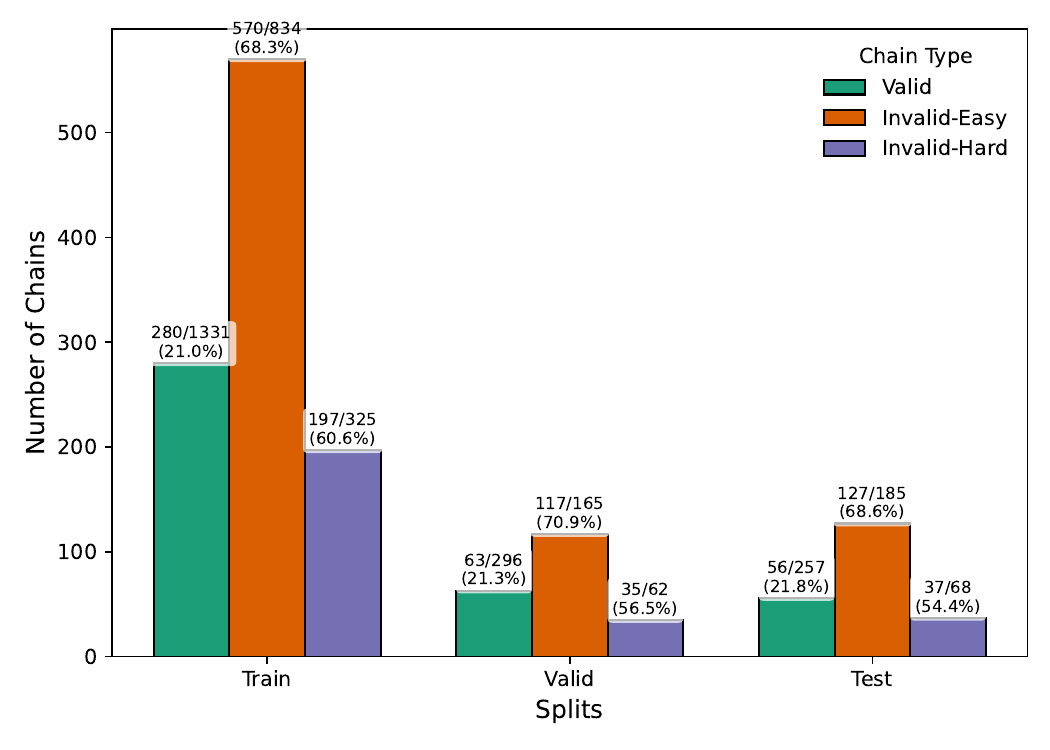}
    \caption{Distribution of valid and invalid chains across splits, also the percentage of chains ending in 2023/2024. For example, the training data consists of 21.0\% valid chains (280 out of 1331) with 2023/2024 paper.}
    \label{fig:dataset_split_summary}
\end{figure}

\section{Supervised Finetuning and Hypothesis generation}\label{sec:experiments}

\begin{table*}
\centering
\small
\resizebox{2\columnwidth}{!}{%
\begin{tabular}{lcccccc}
\toprule
\textbf{Task} & \textbf{Model} & \textbf{Accuracy} & \textbf{Precision} & \textbf{Recall} & \textbf{F1-score} & \textbf{Support} \\
\midrule
\multicolumn{7}{c}{\textbf{Classification Performance}} \\
\midrule
\multirow{2}{*}{\textbf{$\texttt{1-hop}$}} 
    & Phi3-3.8B & 23.41\% & 0.80 & 0.23 & 0.17 & 819 \\ 
    & Llama-3.2 & 35.16\% & 0.79 & 0.35 & 0.35 & 819 \\ 
    & \texttt{HypER}_{Phi3-3.8B} & 72.04\% & \textbf{0.84} & 0.72 & 0.77 & 819 \\
    & \texttt{HypER}_{Llama-3.2} & \textbf{73.87\%} & \textbf{0.84}  &   \textbf{0.74}   &   \textbf{0.78}    &   819 \\
\cmidrule{1-7}
\multirow{2}{*}{\textbf{$\texttt{multi-hop-A}$}} 
    & Phi3-3.8B & 76.86\% & 0.77 & 0.77 & 0.77 & 510 \\
    & Llama-3.2 & 58.24\% & 0.62 & 0.58 & 0.55 & 510 \\ 
    & \texttt{HypER}_{Phi3-3.8B} & \textbf{84.71\%} & \textbf{0.85} & \textbf{0.85} & \textbf{0.85} & 510 \\
    & \texttt{HypER}_{Llama-3.2} & 80.78\% & 0.82   &   0.81  &    0.81   &    510 \\
\cmidrule{1-7}
\multirow{2}{*}{\textbf{$\texttt{multi-hop-C}$}} 
    & Phi3-3.8B & 55.69\% & 0.61 & 0.56 & 0.50 & 510 \\ 
    & Llama-3.2 & 55.10\% & 0.55 & 0.55 & 0.54 & 510 \\ 
    & \texttt{HypER}_{Phi3-3.8B} & 85.66\% & 0.86 & 0.86 & 0.86 & 509 \\
    & \texttt{HypER}_{Llama-3.2} & \textbf{90.39\%} &  \textbf{0.92}   &   \textbf{0.90}    &   \textbf{0.90}    &    510 \\
\midrule
\multicolumn{7}{c}{\textbf{Invalid Node Identification (Invalid Paper ID Matching)}} \\
\multirow{2}{*} & &  & \textbf{Precision} & \textbf{Recall} & \textbf{F1-score} & \textbf{Jaccard Sim.} \\
\midrule
\multirow{2}{*}{\textbf{$\texttt{multi-hop-A}$}} 
    & Phi3-3.8B &  &  0.09 & 0.19 & 0.11 & 0.48 \\
    & Llama-3.2 &  & 0.09 & 0.19 & 0.11 & 0.24 \\ 
    & \texttt{HypER}_{Phi3-3.8B} &  & \textbf{0.30} & \textbf{0.34} & \textbf{0.30} & \textbf{0.65} \\
    & \texttt{HypER}_{Llama-3.2} &  &  0.27   &   0.32    &   0.28    &    0.60 \\
\cmidrule{1-7}
\multirow{2}{*}{\textbf{$\texttt{multi-hop-C}$}} 
    & Phi3-3.8B &  & 0.12 & 0.24 & 0.14 & 0.21 \\
    & Llama-3.2 &  & 0.06 & 0.09 & 0.06 & 0.39 \\
    & \texttt{HypER}_{Phi3-3.8B} &  & \textbf{0.30} & \textbf{0.33} & \textbf{0.30} & \textbf{0.66} \\
    & \texttt{HypER}_{Llama-3.2} &  & \textbf{ 0.30}   &   \textbf{0.33}   &   \textbf{0.30}    &    \textbf{0.66} \\
\midrule
\multicolumn{7}{c}{\textbf{Overall Performance}} \\
\midrule
\textbf{Metric} & Phi3-3.8B & \texttt{HypER}_{Phi3-3.8B} & Llama-3.2 & \texttt{HypER}_{Llama-3.2} &  \texttt{mistral (4 bits)} & \texttt{HypER}_{mistral (4 bits)} \\
\midrule
Average F1-Score 
& 0.468 &
 \textbf{0.616} &
 0.414  &
\textbf{0.614} &
N/A &
 0.43 \\
\bottomrule
\end{tabular}
}
\caption{Comparison of baselines (Phi3-3.8B, LLaMA 3.2) with and without HypER fine-tuning, \texttt{HypER\_*} rows reflect models finetuned with multi-task supervision over the chain dataset. Baseline models are instruction-tuned only, without access to chain structure or intermediate reasoning signals. Full per-class breakdowns for the classification task using \texttt{HypER}\_{Phi3-3.8B} are available in Table~\ref{tab:appendix_classification} of Appendix~\ref{appendix:valid_chain_prediction}.}
\label{tab:rq2_full_results}
\end{table*}

\paragraph{Dataset Splitting} The fine-tuning dataset, consists of $3523$ chains was split into training ($70\%$), validation ($15\%$), and test ($15\%$) sets using a review-ID based grouping strategy to prevent data leakage. Chains were categorized into \textit{valid}, \textit{invalid-easy} and \textit{invalid-hard} categories through randomized review-ID shuffling to have unbiased distribution (see Figure~\ref{fig:dataset_split_summary}). Additional details about data balancing are in the appendix. 

\begin{figure*}[ht]
     \centering
     \includegraphics[width=\textwidth]{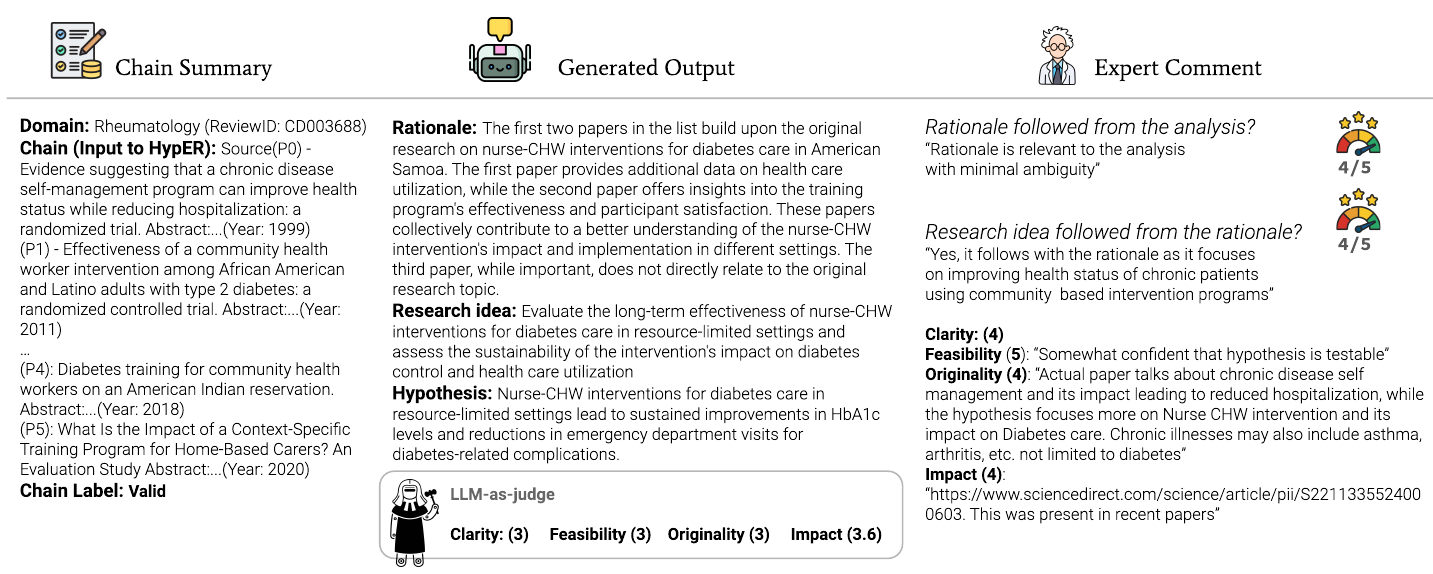}
     \caption{Example of a hypothesis generated by \texttt{HypER}. The output includes the model-generated rationale, research idea, and hypothesis, along with automated (LLM-as-judge) and expert evaluations. Expert reviewers rated the output as relevant, feasible, and clearly connected to prior literature, with moderate novelty and impact. See Appendix~\ref{appendix: expert_analysis} for more details and a contrasting case.}
     \label{fig:sat_expert_example}
 \end{figure*}

\paragraph{Fine-tuning setup for multi-task learning}
To construct a reasoning-driven hypothesis generation model, \texttt{HypER}, we fine-tune SLMs in a multi-task learning setup using the training split of our dataset. 
We consider Phi-3-mini-128k-instruct-3.8B~\citep{abdin2024phi}, instruction-tuned LLaMA‑3.2‑3B model (\texttt{meta-llama/Llama-3.2-3B-Instruct})~\citep{dubey2024llama}, and MistralLite-7B-32K~\footnote{\url{https://huggingface.co/amazon/MistralLite}}, selected for their capability to handle longer context lengths, which is essential for processing extended reasoning chains, to determine the most effective one~\footnote{MistralLite was run with 4-bit quantization (load\_in\_4bit: true). MistralLite was excluded from the final evaluation Table~\ref{tab:rq2_full_results} as it failed to produce outputs in the prompted format most of the time making parsing and processing inconsistent with other models}.
Fine-tuned models consistently outperformed their base counterparts in multi-task learning, with \texttt{HypER}\_Phi3 performing best. Given this, \emph{we use \texttt{HypER}\_Phi3-3.8B and Phi-3-3.8B for subsequent experiments.} We employ Low-Rank Adaptation (LoRA)~\citep{hu2021lora}, a parameter-efficient fine-tuning method, with a rank of 8, a learning rate of $2e-5$, and adapter modules applied to attention layers.
The training dataset is defined as $\mathcal{D}_{train} = \sum_{i=1}^N \{ (\mathcal{C}_i, y_i) \}$, where each reasoning chain $\mathcal{C}_i$ is labeled as valid or invalid. The multi-task setup (Section~\ref{sec:multi-task}) enables reasoning chain classification, invalid node detection, and relevance prediction-leveraging extended context to model fine-grained scientific dependencies. We then evaluate the model's ability to generate hypotheses conditioned on validated chains. Additional compute details are in Appendix~\ref{appendix:compute}.

\paragraph{Metrics}
We evaluate classification performance using accuracy, precision, recall, and F1. For invalid node identification, we report Jaccard similarity~\citep{thada2013comparison} to see the invalid node overlap. Hypothesis quality is assessed in terms of novelty, plausibility, and alignment with literature, following the judging protocols of~\cite{baek2024researchagent} for both LLM-as-a-judge and human judges. Novelty is measured following~\citet{lu2024ai} by iterative literature queries and decides if an idea introduces new insights. We expect that the rationale and hypothesis generated by \texttt{HypER} will focus more on the valid part of a noisy chain and exhibit more coherence than the base model in the presence of noise. Hence, we evaluate the coherence and the groundedness of the generated rationale and hypothesis. Besides human evaluation of the above, we also computed groundedness as faithfulness\footnotetext{\small Faithfulness refers to how well the rationale and hypothesis reflect the content and logic of the input chain, measured by Alignscore.} to the input chain in terms of Alignscore~\citep{zha2023alignscore}.

\section{Experimental Results}
\label{sec:results}

To systematically evaluate our approach, we investigate the following research questions, aligned with the core objectives of reasoning chain validation and hypotheses generation:

\paragraph{RQ1: Can \texttt{HypER} differentiate between valid and invalid reasoning chains?} Hypothesis generation often suffers from noisy information due to weak conceptual links in retrieved literature. Table~\ref{tab:rq2_full_results} shows \texttt{HypER}\_{Phi3-3.8B} (as \texttt{HypER}) significantly improves the Phi3-3.8B base-model (baseline) across all tasks. In one-hop relevance classification, \texttt{HypER} improves F1-score from 17\% to 77\%, indicating its strong ability to capture fine-grained scientific dependencies. For multi-hop chain validation, \texttt{HypER} achieves 85\% ($\uparrow$8) and 86\% ($\uparrow$36) over the Phi3-3.8B base model on both multi-hop chain validation tasks, respectively.
\texttt{HypER} is also much better at identifying incorrect papers in invalid chains, with a Jaccard similarity (overlapping lists) of 0.65 vs. 0.48 by Phi3-3.8B. Overall, \texttt{HypER} improves reasoning chain classification by +22\% F1 over the base model, averaged across two multi-hop validation tasks, making it more effective at scientific reasoning tasks to enhance hypothesis generation with explanation.
\noindent
\textbf{\faSearch~Takeaway:} \texttt{HypER} effectively distinguishes valid from invalid reasoning chains in noisy literature graphs, demonstrating strong performance across multi-tasks.

\begin{table}[htbp!]
\centering
\renewcommand{\arraystretch}{1.2}
\tiny
\begin{tabular}{|c|c|c|c|}
\hline
Model & Chain & Novelty & Explanation groundedness \\
\hline
\multirow{2}{*}{Base-model 1-shot} & valid & 20/30  & 0.305 $\pm$ 0.12 \\
\cline{2-4}
& easy negative & 47/72 &  0.303 $\pm$ 0.11 \\
& hard negative & 20/27 &  0.269 $\pm$ 0.14 \\
\hline
\multirow{2}{*}{HypER 1-shot} & valid & 18/30 & 0.327 $\pm$ 0.14 \\
\cline{2-4}
& easy negative & 44/70 & 0.364 $\pm$ 0.11 \\
& hard negative & 11/26 & 0.324 $\pm$ 0.18 \\
\hline
\end{tabular}
\caption{HypER is better able to ground hypotheses than the base model on the subset of the test-data (2024-target chains).}
\label{tab:metric}
\end{table}

\paragraph{RQ2: Does reasoning chain validation improve the quality of generated hypotheses?} We evaluate this using both automatic and human assessments.
\paragraph{Automatic evaluation:} While there are many existing methods to generate hypotheses, they often lack justification of how the hypothesis was formed. We evaluated novelty and groundedness of rationale on 2024-target chains (Table~\ref{tab:metric}). HypER consistently outperforms base models across valid and noisy chain types, generating better-supported rationales.
For valid chains, \texttt{HypER} achieves a groundedness score of $0.327 \pm 0.14$, compared to $0.305 \pm 0.12$ for the base model, suggesting that reasoning chain validation enhances the model's ability to ground hypotheses in scientific evidence.
Interestingly, for hard negative chains, \texttt{HypER} exhibits significant alignment with valid part of the chain, suggesting that the model can identify and leverage coherent reasoning structures within partially invalid chains. 
However, a PubMedBERT~\citep{gu2021domain} embedding-based similarity analysis between consecutive papers reveals similar semantic overlap for valid and invalid citation-based chains (valid: $0.988 \pm 0.005$, invalid: $0.987 \pm 0.006$). This indicates that semantic similarity alone does not guarantee logical coherence, emphasizing the need for explicit reasoning validation in hypotheses generation. 
\noindent
\textbf{\faSearch~Takeaway:} Validating reasoning chains enhances \texttt{HypER}'s ability to generate hypotheses that are more plausibly grounded in scientific evidence, even in the presence of a noisy reasoning chain ($0.269$-$0.303$ to $0.324$-$0.364$ on negative chains). 
For automated comparisons with larger proprietary models such as GPT-4o, see Appendix~\ref{appendix:auto_eval_gpt4}.

\paragraph{Human Evaluation: } \label{result:human}
We asked 10 medical experts from Upwork (details in appendix~\ref{appendix: expert_analysis}) to evaluate the quality of the analysis, rationale, research idea, and hypothesis generated by \texttt{HypER}\_{Phi-3-3.8B}. We gave five of each type of chains (valid, easy negative, and hard negative) to at least 3 experts and asked whether the model's analysis was correct for every paper in the chain. 
They also answered the following questions on a 5-point Likert scale:
\textbf{1.} whether the rationale followed from the analysis,
\textbf{2.} whether the research idea followed from the rationale, 
\textbf{3.} clarity of hypothesis, 
\textbf{4.} originality (when compared with the articles in the chain) of hypothesis,
\textbf{5.} feasibility of the hypothesis, and 
\textbf{6.} impact of the hypothesis. 
For clarity, originality, feasibility, and impact, we follow the same rubric as the LLM-as-judge \cite{Baek2024ResearchAgentIR}. 
The ratings ($\mu \pm \sigma$) were as follows: (1) rationale consistency 3.47 $\pm$ 0.91, (2) research idea consistency with rationale 3.9 $\pm$ 0.88, (3) hypothesis clarity 3.88 $\pm$ 0.47, (4) originality 3.21 $\pm$ 0.5, (5) feasibility 4.22 $\pm$ 1.2, and (6) impact 3.69 $\pm$ 0.54. Full rating breakdowns and examples are provided in Appendix~\ref{appendix: expert_analysis}. As detailed in Appendix~\ref{appendix:llm_human_correlation}, we found moderate human-LLM agreement on clarity ($r=0.53$, $\rho=0.57$) and impact ($r=0.57$, $\rho=0.51$), but weaker alignment on originality and feasibility ($\rho = -0.08$), highlighting that LLMs may overvalue fluency over scientific grounding.
\noindent
\textbf{\faSearch~Takeaway:} Our results (Figure~\ref{fig:sat_expert_example}) indicate that \texttt{HypER}\_{Phi-3-3.8B} generates scientifically grounded novel hypotheses rather than arbitrary hypotheses, making it a more reliable tool for literature-based discovery.

\section{Related Work}
Providing structured explanation for hypotheses has been emphasized in AI-driven drug discovery~\citep{sudhahar2024experimentally}. In experimental sciences, \citet{boiko2023autonomous} integrate GPT-4 with external tools such as web and document search, while~\citet{abdel2024scientific} leverage hallucinations to hypothesize novel pairs of FDA approved cancer drugs in breast cancer treatment, arguing that the validity can ultimately be experimentally verified. In the social sciences,~\citet{yang2023large} propose a multi-module framework for feedback exploration. DiscoveryBench~\cite{majumder2024discoverybench} formalizes hypotheses as semantic trees, though focused on data-driven rather than literature-based discovery. Unlike these methods, \texttt{HypER} incorporates explicit reasoning validation by validating dependencies in literature graphs, ensuring that hypotheses are derived from logically coherent and evidence-backed research trajectories. Systems like SCIMON~\citep{wang2023learning} and ResearchAgent~\citep{baek2024researchagent} support LLM-based ideation but do not validate scientific dependencies: SCIMON targets novelty without structural justification, and ResearchAgent relies on agent-based refinement with loosely connected papers, but neither provides structured evidence tracing how a hypothesis emerges. As shown in Section~\ref{sec:results}, even invalid citation chains exhibit high semantic similarity to valid ones ($\sim$0.98), highlighting that semantic similarity alone fails to capture scientific reasoning -a distinction HypER explicitly models (elaborated in Appendix~\ref{appendix:Related_Work}).

\section{Conclusion}\label{sec:conclusion}
\texttt{HypER} introduces fine-grained reasoning validation for literature-based hypothesis generation, ensuring that generated hypotheses are not only plausible but also scientifically grounded. 
Unlike prior methods that rely on surface-level retrieval, \texttt{HypER} constructs and validates structured reasoning chains, filtering out misleading connections and reinforcing logical coherence. Our results show that \texttt{HypER}\_{Phi-3-3.8B} significantly improves AI-supported hypothesis generation, making research ideation more structured and evidence-driven. 
This has broad implications--accelerating research, helping scientists navigate complex literature, and pushing AI toward more structured scientific reasoning.

\section{Limitations}
Our approach construct chains using abstracts to fit within model context limits and to circumvent the scarcity of open-access full-text medical literature. However, this abstract-based method does not fully capture the real-world scientific discovery process, where researchers have to read them in entirety, after shortlisting the relevant articles. 
Additionally, the necessary rigor of scientific literature review process limited our human evaluation process. Due to the complexity of assessing reasoning chains, we conducted evaluations on a limited sample size. In particular, our correlation analysis between expert and LLM-as-judge ratings is based on just 15 examples, which may not capture the full variability in evaluation behavior. While trends are informative, these results should be interpreted with caution and validated on larger datasets in future work.
However, a challenging task as this would require more elaborate and pragmatic evaluation.

Furthermore, our fine-tuned model inherits certain weaknesses from the base model such as copying from few-shot example, which may have limited the model performance in some generated instances.
While \texttt{HypER} is effective at filtering meaningful reasoning paths from misleading ones, it is not explicitly designed to optimize for novelty. A future extension of this work could focus on fine-tuning \texttt{HypER} to better balance plausibility and novelty in hypothesis generation.

We did not include full-scale comparisons using proprietary models such as GPT-4o or fine-tuning experiments with larger LLMs. Our objective is to train small, instruction-tuned models that are openly accessible, reproducibly fine-tuned, and efficient to deploy. While constructing the reasoning chains required costly citation graph traversal and large-model queries, we distilled a smaller model (Phi-3-mini) from a validated LLM, enabling efficient inference while preserving reasoning quality. 
To benchmark HypER’s performance, we conducted an automated output-level evaluation against GPT-4o, showing that HypER achieves comparable scores in originality (3.01 vs. 3.00) and significance (3.37 vs. 3.84). 
However, since a comprehensive human evaluation would be necessary for a fair and rigorous comparison, we leave that to future work. See Appendix~\ref{appendix:auto_eval_LLM} for further details.

\section{Ethics Statement}
We honor the Code of Ethics. No personally identifiable information is collected or used in this work. The human evaluators were hired from Upwork using a detailed job post. We had Institutional Review Board (IRB) approval for obtaining written consent from our human evaluators. We shared an example task sheet with complete instructions during the recruitment. The evaluators were duly compensated based on minimum wage in the respective countries and always above their quotation.

\section*{Acknowledgments}
This work is partly funded by the Swiss National Science Foundation through project \href{https://data.snf.ch/grants/grant/184994}{``CrowdAlytics''} (contract no\ 184994) and project \href{https://data.snf.ch/grants/grant/205975}{``Digital Deliberative Democracy''} (contract no\ 205975).

\bibliography{custom}

\begin{thebibliography}{38}
\providecommand{\natexlab}[1]{#1}

\bibitem[{Abdel-Rehim et~al.(2024)Abdel-Rehim, Zenil, Orhobor, Fisher, Collins,
  Bourne, Fearnley, Tate, Smith, Soldatova et~al.}]{abdel2024scientific}
Abbi Abdel-Rehim, Hector Zenil, Oghenejokpeme Orhobor, Marie Fisher, Ross~J
  Collins, Elizabeth Bourne, Gareth~W Fearnley, Emma Tate, Holly~X Smith,
  Larisa~N Soldatova, et~al. 2024.
\newblock Scientific hypothesis generation by a large language model:
  Laboratory validation in breast cancer treatment.
\newblock \emph{arXiv e-prints}, pages arXiv--2405.

\bibitem[{Abdin et~al.(2024)Abdin, Aneja, Awadalla, Awadallah, Awan, Bach,
  Bahree, Bakhtiari, Bao, Behl et~al.}]{abdin2024phi}
Marah Abdin, Jyoti Aneja, Hany Awadalla, Ahmed Awadallah, Ammar~Ahmad Awan,
  Nguyen Bach, Amit Bahree, Arash Bakhtiari, Jianmin Bao, Harkirat Behl, et~al.
  2024.
\newblock Phi-3 technical report: A highly capable language model locally on
  your phone.
\newblock \emph{arXiv preprint arXiv:2404.14219}.

\bibitem[{Baek et~al.(2024{\natexlab{a}})Baek, Jauhar, Cucerzan, and
  Hwang}]{baek2024researchagent}
Jinheon Baek, Sujay~Kumar Jauhar, Silviu Cucerzan, and Sung~Ju Hwang.
  2024{\natexlab{a}}.
\newblock Researchagent: Iterative research idea generation over scientific
  literature with large language models.
\newblock \emph{arXiv preprint arXiv:2404.07738}.

\bibitem[{Baek et~al.(2024{\natexlab{b}})Baek, Jauhar, Cucerzan, and
  Hwang}]{Baek2024ResearchAgentIR}
Jinheon Baek, Sujay~Kumar Jauhar, Silviu Cucerzan, and Sung~Ju Hwang.
  2024{\natexlab{b}}.
\newblock \href {https://api.semanticscholar.org/CorpusID:269042844}
  {Researchagent: Iterative research idea generation over scientific literature
  with large language models}.
\newblock \emph{ArXiv}, abs/2404.07738.

\bibitem[{Bichindaritz et~al.(1998)Bichindaritz, Kansu, and
  Sullivan}]{bichindaritz1998case}
Isabelle Bichindaritz, Emin Kansu, and Keith~M Sullivan. 1998.
\newblock Case-based reasoning in care-partner: Gathering evidence for
  evidence-based medical practice.
\newblock In \emph{European workshop on advances in case-based reasoning},
  pages 334--345. Springer.

\bibitem[{Boiko et~al.(2023)Boiko, MacKnight, Kline, and
  Gomes}]{boiko2023autonomous}
Daniil~A Boiko, Robert MacKnight, Ben Kline, and Gabe Gomes. 2023.
\newblock Autonomous chemical research with large language models.
\newblock \emph{Nature}, 624(7992):570--578.

\bibitem[{Dubey et~al.(2024)Dubey, Jauhri, Pandey, Kadian, Al-Dahle, Letman,
  Mathur, Schelten, Yang, Fan et~al.}]{dubey2024llama}
Abhimanyu Dubey, Abhinav Jauhri, Abhinav Pandey, Abhishek Kadian, Ahmad
  Al-Dahle, Aiesha Letman, Akhil Mathur, Alan Schelten, Amy Yang, Angela Fan,
  et~al. 2024.
\newblock The llama 3 herd of models.
\newblock \emph{arXiv preprint arXiv:2407.21783}.

\bibitem[{Ghafarollahi and Buehler(2024)}]{ghafarollahi2024sciagents}
Alireza Ghafarollahi and Markus~J Buehler. 2024.
\newblock Sciagents: Automating scientific discovery through bioinspired
  multi-agent intelligent graph reasoning.
\newblock \emph{Advanced Materials}, page 2413523.

\bibitem[{Gu and Krenn(2024)}]{gu2024generation}
Xuemei Gu and Mario Krenn. 2024.
\newblock Generation and human-expert evaluation of interesting research ideas
  using knowledge graphs and large language models.
\newblock \emph{arXiv preprint arXiv:2405.17044}.

\bibitem[{Gu et~al.(2021)Gu, Tinn, Cheng, Lucas, Usuyama, Liu, Naumann, Gao,
  and Poon}]{gu2021domain}
Yu~Gu, Robert Tinn, Hao Cheng, Michael Lucas, Naoto Usuyama, Xiaodong Liu,
  Tristan Naumann, Jianfeng Gao, and Hoifung Poon. 2021.
\newblock Domain-specific language model pretraining for biomedical natural
  language processing.
\newblock \emph{ACM Transactions on Computing for Healthcare (HEALTH)},
  3(1):1--23.

\bibitem[{Hu et~al.(2021)Hu, Shen, Wallis, Allen-Zhu, Li, Wang, Wang, and
  Chen}]{hu2021lora}
Edward~J Hu, Yelong Shen, Phillip Wallis, Zeyuan Allen-Zhu, Yuanzhi Li, Shean
  Wang, Lu~Wang, and Weizhu Chen. 2021.
\newblock Lora: Low-rank adaptation of large language models.
\newblock \emph{arXiv preprint arXiv:2106.09685}.

\bibitem[{Jing et~al.(2024)Jing, Cimino, Patel, Zhou, Shubrook, Liu, and
  De~Lacalle}]{jing2024data}
Xia Jing, James~J Cimino, Vimla~L Patel, Yuchun Zhou, Jay~H Shubrook, Chang
  Liu, and Sonsoles De~Lacalle. 2024.
\newblock Data-driven hypothesis generation in clinical research: What we
  learned from a human subject study?
\newblock \emph{Medical Research Archives}, 12(2).

\bibitem[{Karunarathna et~al.(2024)Karunarathna, Gunasena, Hapuarachchi, and
  Gunathilake}]{karunarathna2024evolution}
Indunil Karunarathna, P~Gunasena, T~Hapuarachchi, and S~Gunathilake. 2024.
\newblock The evolution of hypotheses in scientific literature: A review of
  impact and reach.

\bibitem[{King et~al.(2009)King, Rowland, Oliver, Young, Aubrey, Byrne,
  Liakata, Markham, Pir, Soldatova, Sparkes, Whelan, and
  Clare}]{doi:10.1126/science.1165620}
Ross~D. King, Jem Rowland, Stephen~G. Oliver, Michael Young, Wayne Aubrey, Emma
  Byrne, Maria Liakata, Magdalena Markham, Pinar Pir, Larisa~N. Soldatova,
  Andrew Sparkes, Kenneth~E. Whelan, and Amanda Clare. 2009.
\newblock \href {https://doi.org/10.1126/science.1165620} {The automation of
  science}.
\newblock \emph{Science}, 324(5923):85--89.

\bibitem[{Kumar et~al.(2024)Kumar, Ghosal, Goyal, and Ekbal}]{kumar2024can}
Sandeep Kumar, Tirthankar Ghosal, Vinayak Goyal, and Asif Ekbal. 2024.
\newblock Can large language models unlock novel scientific research ideas?
\newblock \emph{arXiv preprint arXiv:2409.06185}.

\bibitem[{Li et~al.(2024)Li, Xu, Guo, Zhao, Li, Yuan, Zhang, Jiang, Xin, Dang
  et~al.}]{li2024chain}
Long Li, Weiwen Xu, Jiayan Guo, Ruochen Zhao, Xingxuan Li, Yuqian Yuan, Boqiang
  Zhang, Yuming Jiang, Yifei Xin, Ronghao Dang, et~al. 2024.
\newblock Chain of ideas: Revolutionizing research via novel idea development
  with llm agents.
\newblock \emph{arXiv preprint arXiv:2410.13185}.

\bibitem[{Li and Ouyang(2024)}]{li-ouyang-2024-related}
Xiangci Li and Jessica Ouyang. 2024.
\newblock \href {https://doi.org/10.18653/v1/2024.emnlp-main.767} {Related work
  and citation text generation: A survey}.
\newblock In \emph{Proceedings of the 2024 Conference on Empirical Methods in
  Natural Language Processing}, pages 13846--13864, Miami, Florida, USA.
  Association for Computational Linguistics.

\bibitem[{Lu et~al.(2024)Lu, Lu, Lange, Foerster, Clune, and Ha}]{lu2024ai}
Chris Lu, Cong Lu, Robert~Tjarko Lange, Jakob Foerster, Jeff Clune, and David
  Ha. 2024.
\newblock The ai scientist: Towards fully automated open-ended scientific
  discovery.
\newblock \emph{arXiv preprint arXiv:2408.06292}.

\bibitem[{Majumder et~al.(2024)Majumder, Surana, Agarwal, Mishra, Meena,
  Prakhar, Vora, Khot, Sabharwal, and Clark}]{majumder2024discoverybench}
Bodhisattwa~Prasad Majumder, Harshit Surana, Dhruv Agarwal, Bhavana~Dalvi
  Mishra, Abhijeetsingh Meena, Aryan Prakhar, Tirth Vora, Tushar Khot, Ashish
  Sabharwal, and Peter Clark. 2024.
\newblock Discoverybench: Towards data-driven discovery with large language
  models.
\newblock \emph{arXiv preprint arXiv:2407.01725}.

\bibitem[{Nadkarni et~al.()Nadkarni, Wadden, Beltagy, Smith, Hajishirzi, and
  Hope}]{nadkarni2021scientific}
Rahul Nadkarni, David Wadden, Iz~Beltagy, Noah Smith, Hannaneh Hajishirzi, and
  Tom Hope.
\newblock Scientific language models for biomedical knowledge base completion:
  An empirical study.
\newblock In \emph{NeurIPS 2021 AI for Science Workshop}.

\bibitem[{Nigam et~al.(2024)Nigam, Patwardhan, Vig, and
  Shroff}]{nigam2024interactive}
Harshit Nigam, Manasi Patwardhan, Lovekesh Vig, and Gautam Shroff. 2024.
\newblock An interactive co-pilot for accelerated research ideation.
\newblock In \emph{Proceedings of the Third Workshop on Bridging
  Human--Computer Interaction and Natural Language Processing}, pages 60--73.

\bibitem[{Pu et~al.(2024)Pu, Feng, Grossman, Hope, Mishra, Latzke, Bragg,
  Chang, and Siangliulue}]{pu2024ideasynth}
Kevin Pu, KJ~Feng, Tovi Grossman, Tom Hope, Bhavana~Dalvi Mishra, Matt Latzke,
  Jonathan Bragg, Joseph~Chee Chang, and Pao Siangliulue. 2024.
\newblock Ideasynth: Iterative research idea development through evolving and
  composing idea facets with literature-grounded feedback.
\newblock \emph{arXiv preprint arXiv:2410.04025}.

\bibitem[{Qi et~al.(2024)Qi, Zhang, Tian, Li, Chen, Zeng, Hua, Jinfang, and
  Zhou}]{qi2024large}
Biqing Qi, Kaiyan Zhang, Kai Tian, Haoxiang Li, Zhang-Ren Chen, Sihang Zeng,
  Ermo Hua, Hu~Jinfang, and Bowen Zhou. 2024.
\newblock Large language models as biomedical hypothesis generators: a
  comprehensive evaluation.
\newblock \emph{arXiv preprint arXiv:2407.08940}.

\bibitem[{Si et~al.(2024)Si, Yang, and Hashimoto}]{si2024can}
Chenglei Si, Diyi Yang, and Tatsunori Hashimoto. 2024.
\newblock Can llms generate novel research ideas? a large-scale human study
  with 100+ nlp researchers.
\newblock \emph{arXiv preprint arXiv:2409.04109}.

\bibitem[{SU et~al.(2023)SU, Kasai, Wu, Shi, Wang, Xin, Zhang, Ostendorf,
  Zettlemoyer, Smith, and Yu}]{su2023selective}
Hongjin SU, Jungo Kasai, Chen~Henry Wu, Weijia Shi, Tianlu Wang, Jiayi Xin, Rui
  Zhang, Mari Ostendorf, Luke Zettlemoyer, Noah~A. Smith, and Tao Yu. 2023.
\newblock \href {https://openreview.net/forum?id=qY1hlv7gwg} {Selective
  annotation makes language models better few-shot learners}.
\newblock In \emph{The Eleventh International Conference on Learning
  Representations}.

\bibitem[{Sudhahar et~al.(2024)Sudhahar, Ozer, Chang, Chadwick, O’Donovan,
  Campbell, Tulip, Thompson, and Roberts}]{sudhahar2024experimentally}
Saatviga Sudhahar, Bugra Ozer, Jiakang Chang, Wayne Chadwick, Daniel
  O’Donovan, Aoife Campbell, Emma Tulip, Neil Thompson, and Ian Roberts.
  2024.
\newblock An experimentally validated approach to automated biological evidence
  generation in drug discovery using knowledge graphs.
\newblock \emph{Nature Communications}, 15(1):5703.

\bibitem[{Swanson(1986)}]{swanson1986undiscovered}
Don~R Swanson. 1986.
\newblock Undiscovered public knowledge.
\newblock \emph{The Library Quarterly}, 56(2):103--118.

\bibitem[{Sybrandt et~al.(2020)Sybrandt, Tyagin, Shtutman, and
  Safro}]{sybrandt2020agatha}
Justin Sybrandt, Ilya Tyagin, Michael Shtutman, and Ilya Safro. 2020.
\newblock Agatha: automatic graph mining and transformer based hypothesis
  generation approach.
\newblock In \emph{Proceedings of the 29th ACM International Conference on
  Information \& Knowledge Management}, pages 2757--2764.

\bibitem[{Thada and Jaglan(2013)}]{thada2013comparison}
Vikas Thada and Vivek Jaglan. 2013.
\newblock Comparison of jaccard, dice, cosine similarity coefficient to find
  best fitness value for web retrieved documents using genetic algorithm.
\newblock \emph{International Journal of Innovations in Engineering and
  Technology}, 2(4):202--205.

\bibitem[{Thilakaratne et~al.(2019)Thilakaratne, Falkner, and
  Atapattu}]{thilakaratne2019systematic}
Menasha Thilakaratne, Katrina Falkner, and Thushari Atapattu. 2019.
\newblock A systematic review on literature-based discovery: general overview,
  methodology, \& statistical analysis.
\newblock \emph{ACM Computing Surveys (CSUR)}, 52(6):1--34.

\bibitem[{Wallace et~al.(2021)Wallace, Saha, Soboczenski, and
  Marshall}]{AMIA-summarization-2021}
Byron~C. Wallace, Sayantan Saha, Frank Soboczenski, and Iain~J. Marshall. 2021.
\newblock {Generating (Factual?) Narrative Summaries of RCTs: Experiments with
  Neural Multi-Document Summarization}.
\newblock In \emph{{Proceedings of AMIA Informatics Summit}}.

\bibitem[{Wang et~al.(2023{\natexlab{a}})Wang, Downey, Ji, and
  Hope}]{wang2023scimon}
Qingyun Wang, Doug Downey, Heng Ji, and Tom Hope. 2023{\natexlab{a}}.
\newblock Scimon: Scientific inspiration machines optimized for novelty.
\newblock \emph{arXiv preprint arXiv:2305.14259}.

\bibitem[{Wang et~al.(2023{\natexlab{b}})Wang, Downey, Ji, and
  Hope}]{wang2023learning}
Qingyun Wang, Doug Downey, Heng Ji, and Tom Hope. 2023{\natexlab{b}}.
\newblock Scimon: Scientific inspiration machines optimized for novelty.
\newblock \emph{arXiv preprint arXiv:2305.14259}.

\bibitem[{Xu et~al.(2023)Xu, Sheng, Xue, Fu, Wang, and Zhou}]{xu2023exploring}
Yi~Xu, Shuqian Sheng, Bo~Xue, Luoyi Fu, Xinbing Wang, and Chenghu Zhou. 2023.
\newblock Exploring and verbalizing academic ideas by concept co-occurrence.
\newblock In \emph{Proceedings of the 61st Annual Meeting of the Association
  for Computational Linguistics (Volume 1: Long Papers)}, pages 13001--13027.

\bibitem[{Xun et~al.(2017)Xun, Jha, Gopalakrishnan, Li, and
  Zhang}]{xun2017generating}
Guangxu Xun, Kishlay Jha, Vishrawas Gopalakrishnan, Yaliang Li, and Aidong
  Zhang. 2017.
\newblock Generating medical hypotheses based on evolutionary medical concepts.
\newblock In \emph{2017 IEEE International conference on data mining (ICDM)},
  pages 535--544. IEEE.

\bibitem[{Yang et~al.(2019)Yang, Steinfeld, and
  Zimmerman}]{DBLP:journals/corr/abs-1904-09612}
Qian Yang, Aaron Steinfeld, and John Zimmerman. 2019.
\newblock \href {https://arxiv.org/abs/1904.09612} {Unremarkable {AI:} fitting
  intelligent decision support into critical, clinical decision-making
  processes}.
\newblock \emph{CoRR}, abs/1904.09612.

\bibitem[{Yang et~al.(2023)Yang, Du, Li, Zheng, Poria, and
  Cambria}]{yang2023large}
Zonglin Yang, Xinya Du, Junxian Li, Jie Zheng, Soujanya Poria, and Erik
  Cambria. 2023.
\newblock Large language models for automated open-domain scientific hypotheses
  discovery.
\newblock \emph{arXiv preprint arXiv:2309.02726}.

\bibitem[{Zha et~al.(2023)Zha, Yang, Li, and Hu}]{zha2023alignscore}
Yuheng Zha, Yichi Yang, Ruichen Li, and Zhiting Hu. 2023.
\newblock Alignscore: Evaluating factual consistency with a unified alignment
  function.
\newblock In \emph{Proceedings of the 61st Annual Meeting of the Association
  for Computational Linguistics (Volume 1: Long Papers)}, pages 11328--11348.

\end{thebibliography}

\appendix

\section{Additional Related Work}\label{appendix:Related_Work}

\begin{table*}[t]
\centering
\small
\begin{tabular}{lp{2.2cm}p{1.8cm}p{2.6cm}p{2.2cm}}
\toprule
\textbf{System} & \textbf{Citation Data} & \textbf{Reasoning Validation} & \textbf{Hypothesis Output} & \textbf{Model Size} \\
\midrule
\textsc{SCIMON}  & \cmark & \xmark & \cmark & Large \\
\textsc{ResearchAgent} & \cmark & \xmark & \cmark & Large \\
~\cite{kumar2024can} & \xmark & \xmark & \cmark & Large  \\
\textsc{HypER (Ours)} & \cmark & \cmark & \cmark & Small (3.8B) \\
\bottomrule
\end{tabular}
\caption{Comparison of literature-based hypothesis generation systems. HypER uniquely combines validated citation reasoning with multi-task fine-tuning to produce grounded hypotheses in small models.}
\label{tab:related-comparison}
\end{table*}

\subsection{Literature-based scientific discovery}
Early approaches to hypothesis generation often focused on linking concepts from distinct parts of the literature to generate new hypotheses. A seminal example of this is Swanson's ABC model, which identified hidden connections between seemingly unrelated scientific papers~\cite{swanson1986undiscovered}. Recent advancements~\cite{sybrandt2020agatha,nadkarni2021scientific,xu2023exploring} include methods like scientific knowledge edge link prediction~\cite{nadkarni2021scientific}, which connects concepts in scientific texts. These advancements leverage sophisticated systems to analyze and predict new relationships within the literature. For instance, AGATHA, a deep-learning hypothesis generation system~\cite{sybrandt2020agatha}, introduces data-driven insights to rank plausible term-pairs among entity sets in the discovery process and achieving high recommendation scores in various biomedical sub-domains.~\cite{xu2023exploring} used temporal link prediction and text generation to verbalize a new idea. Unlike these methods, \texttt{HypER} ensures hypotheses are logically coherent by explicitly validating scientific dependencies in literature graphs.

\subsection{Comparison with related ideation systems}
Our goal is to generate hypotheses \textit{explicitly connected to the literature via reasoning chains}; SCIMON's goal~\citep{wang2023scimon} is to generate hypotheses optimized for novelty (literature-inspired, but with no formal connection to the literature). ResearchAgent~\citep{baek2024researchagent} similarly generates hypotheses through agent interactions, but again without a structure explaining how the hypothesis follows from specific papers. Thus, neither system would be able to distinguish valid and invalid reasoning chains, but that would not be a fair comparison as that is not what they were designed for. While SCIMON and ResearchAgent rely on semantic similarity (e.g., abstracts and citations) to guide idea generation, they do not validate logical reasoning between connected ideas. This distinction is critical because, as reported in section~\ref{sec:results}, high semantic similarity ($\sim$0.98) between consecutive papers (both valid and invalid) highlights that semantic overlap alone is insufficient to ensure valid reasoning. Another approach by~\cite{kumar2024can} generates ideas using full-text inputs, while \texttt{HypER} focuses on validating citation-based reasoning chains with a distilled model, making direct comparison less applicable. A comparative summary of these systems is provided in Table~\ref{tab:related-comparison}, highlighting the differences in reasoning validation, citation use, and model size.

\subsection{Additional Discussion}
\label{appendix:discussion}
\paragraph{Implications for Related Work Generation} Recent work~\citep{li-ouyang-2024-related} has highlighted that many related work generation (RWG) models fail to correctly order and group citations, often placing unrelated works together and reducing readability. Our reasoning chain validation approach can help address this by ensuring that only logically connected papers are grouped, promoting more coherent and interpretable citation structures. Moreover, RWG models typically assume that a set of relevant citations is provided, which is not always realistic. In contrast, \texttt{HypER}'s ability to identify intermediate reasoning chains opens up the possibility of retrieval-augmented citation discovery - suggesting missing yet relevant works and improving the completeness of the related work section. We believe this points to a promising future direction, where reasoning-based approaches can strengthen automatic RWG systems by grounding citation structure in validated scientific dependencies.

\section{Summary of Valid Reasoning Chains}\label{chain_stats}
\begin{figure}[h!]
\centering
\includegraphics[width=\columnwidth]{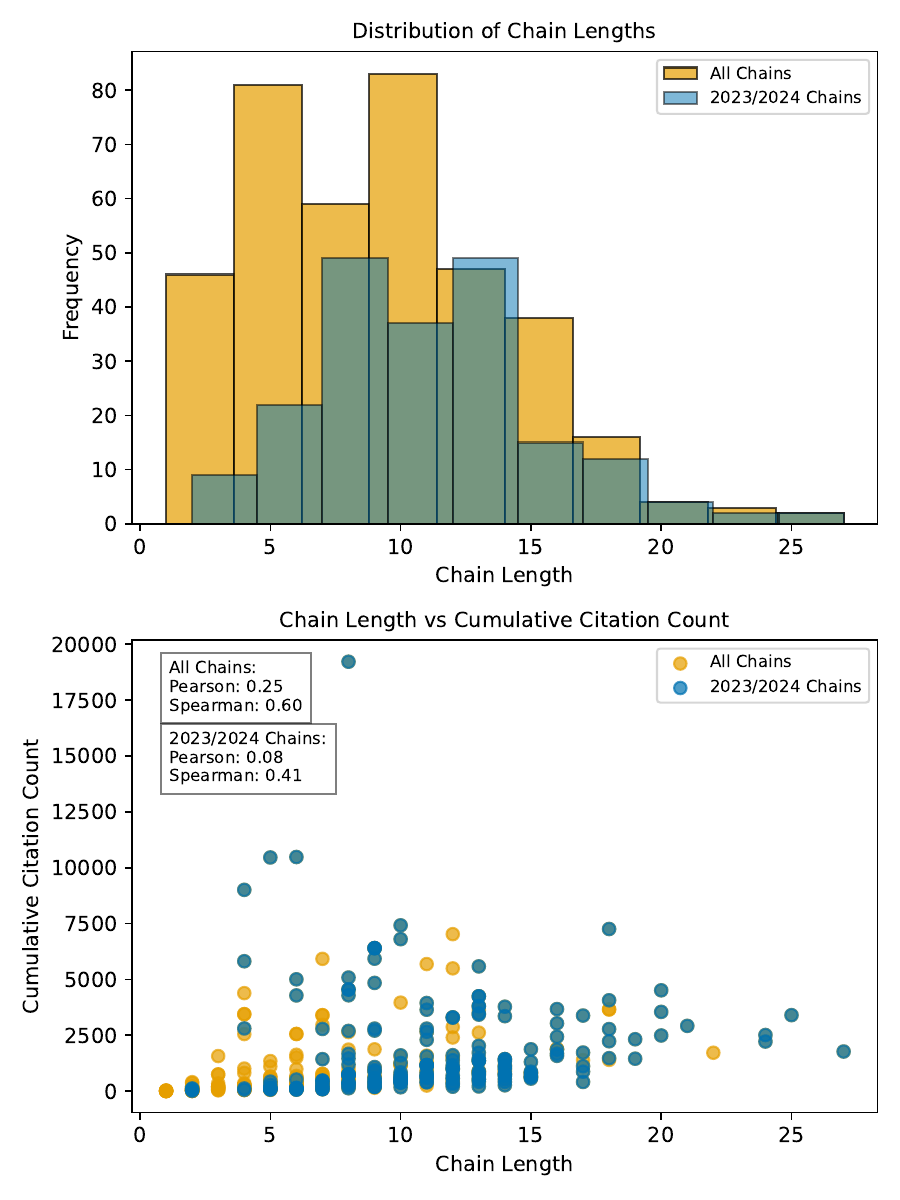}
\caption{(Top) Distribution of chain lengths for all chains (orange) and 2023/2024 chains (blue). (Bottom) Chain length vs. cumulative citation count, illustrating the relationship between length and citation impact across all chains. 
}
\label{fig:chain_analysis}
\end{figure}

We constructed a total of \textit{379 reasoning chains}, each representing a structured progression of ideas connecting a source paper to a hypothesis. The lengths of these chains varied from 1 to 27 papers, with an average length of \textbf{9.04} ($\sigma = 4.76$), reflecting diverse complexities in reasoning paths. Approximately \textbf{53.03\% (201 chains)} concluded with papers published in 2023 or 2024. The cumulative citation counts of the 379 reasoning chains ranged from 0 to 19'219, with a median of \textbf{680}. Chains concluding in 2023/2024 exhibited slightly higher structural complexity, as indicated by their greater average length of \textbf{10.94} ($\sigma = 4.39$) and a median citation count of \textbf{869.0}, highlighting their temporal depth and influence in capturing recent research trends. A detailed analysis of chain length distributions and their relationship to cumulative citation counts is provided in Figure ~\ref{fig:chain_analysis}. Figure~\ref{fig:chain_analysis} illustrates the  relationship between length and citation impact across all chains. Longer chains tend to incorporate more highly cited papers, the relationship is not strictly proportional or linear, as other factors influence citation counts.

We computed the fraction of papers with a relevance score of 2 (excluding the first paper) for each reasoning chain. The mean relevance fraction was $0.71$, with $56.20$\% ($213$ chains) exceeding this threshold. Notably, $19.26$\% ($73$ chains) were fully relevant (fraction = $1.0$), while $12.92$\% ($49$ chains) had fractions below $0.5$. These results indicate that most reasoning chains have a high proportion of relevant papers, demonstrating strong connections.

\section{Sampling reviews and PubMed abstracts from RCT dataset.}\label{appendix:inputRCT}
Our data source consists of $\approx$4.5K systematic reviews of randomized control trials (RCT), each linked to a set of abstracts and spanning several subdisciplines of medicine. To encourage novel interdisciplinary discovery and for targeted expert evaluation, we sampled reviews from 4 distinct yet interacting subdisciplines of medicine. We used a sampling strategy called vote-k~\cite{su2023selective}. Vote-k prioritizes instances with more neighbors (votes) while maintaining diversity by penalizing selections too similar to already chosen samples. This ensures balanced representation from each domain. Using the reviews and associated abstracts from the selected sub-disciplines, we will construct the input chains, which will serve as the basis for our experiments.
The following are the selected subdisciplines: (1)\textit{ Endocrinology:} The study of hormones and glands that control things like growth, metabolism, and reproduction.
(2) \textit{Cardiology:} The study of the heart and blood vessels, focusing on heart diseases and related conditions. (3) \textit{Rheumatology:} The study of joint, muscle, and autoimmune diseases like arthritis and lupus. (4) \textit{Gastroenterology and Hepatology:} The study of the digestive system. Gastroenterology covers the stomach and intestines, while Hepatology focuses on the liver and related organs.

\section{Relevancy scoring using Llama-3.1-70B}\label{appendix:llamaScoring}
\begin{figure}[t]
    \centering
    \includegraphics[width=\linewidth]{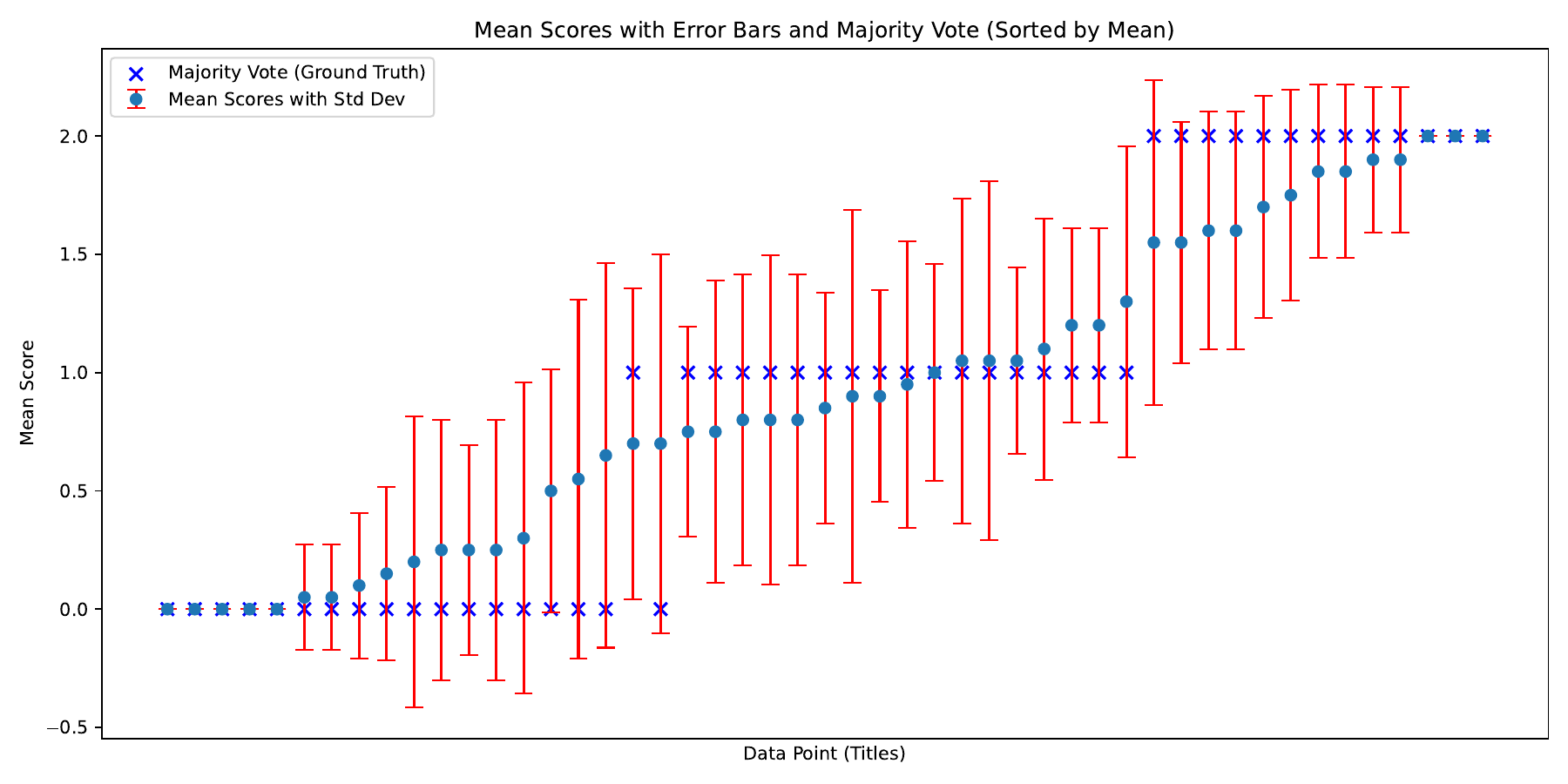}
    \caption{Mean scores with standard deviation error bars across random seeds and their alignment with majority vote results.}
    \label{fig:mean_scores_error_bars}
\end{figure}

The analysis of 50 papers using \texttt{Llama-3.1-70B} across 20 random seeds reveals moderate consistency in relevancy scoring. The overall mean deviation from the majority vote was $0.269$, with a standard deviation of $0.197$, indicating variability in individual runs. Fleiss' Kappa score of $0.458$ suggests moderate agreement among the relevancy scores. Figure~\ref{fig:mean_scores_error_bars} highlights the variability of scores across random seeds and their alignment with the majority vote results.

The LLM's predictions show varying degrees of agreement with individual experts: Cohen's Kappa values of 0.521, 0.382, and 0.384, with percentage agreements of 68.63\%, 58.82\%, and 60.78\%, respectively. When compared against the majority vote of human experts, the LLM achieved a Cohen's Kappa of 0.459 and a percentage agreement of 64.71\%, indicating moderate agreement. Expert annotations also revealed variability in inter-annotator agreement, with Cohen's Kappa scores of 0.472 between expert1 and expert2, 0.382 between expert1 and expert3, and 0.251 between expert2 and expert3. On average, inter-annotator agreement reached a Cohen's Kappa of $0.368 \pm 0.091$

\paragraph{Dataset quality under moderate agreement} To assess dataset quality, we examined alignment between LLaMA-3.1-70B-Instruct and expert reasoning. Given the subjectivity of scientific relevance, moderate agreement is expected. Since our chain construction approach (Section~\ref{chain_construction}) selects the most relevant paper (score = 2) at each hop, falling back to less relevant (score = 1) only if none are available, we examined false positives where the model rated a paper as 2 but experts rated it 0. These disagreements between the model and experts were much lower for relevant papers, and the value is 7.8\%, 2.0\%, and 0\% respectively for expert1, expert2, and expert3 -- suggesting the model rarely selects clearly irrelevant papers as top candidates. This supports its use as a practical, context-aware reasoning proxy for large-scale relevance judgments in scientific chain construction. 

\section{Training specifics} \label{appendix:compute}
To achieve a balanced dataset, longer valid chains were split into overlapping sub-chains of up to 5 papers while retaining the original chains. 
The final count of $3,523$ chains reflects this balancing process rather than the number of initially constructed reasoning chains. Figure~\ref{fig:dataset_split_summary} illustrates the distribution of valid and invalid chains across train, validation and test splits and the proportion of chains ending in 2023 or 2024. 

We employ Low-Rank Adaptation (LoRA)~\citep{hu2021lora}, a parameter-efficient fine-tuning method, with a rank of 8, a learning rate of $2e-5$, and adapter modules applied to attention layers. We used the Axolotl framework\footnote{\url{https://github.com/OpenAccess-AI-Collective/axolotl}} for managing LoRA fine-tuning pipelines and reproducibility across small LLMs. Each reasoning chain is tokenized to include paper titles, abstracts, and extracted target hypotheses (optionally), forming structured input sequences. All experiments were conducted on a high-performance computing system equipped with 8 NVIDIA GeForce RTX 4090 GPUs.

\section{Reasoning path validity prediction task}\label{appendix:valid_chain_prediction}

Table~\ref{tab:appendix_classification} shows the break-down of the performance by \texttt{HypER}\_{Phi-3} on different classes corresponding to each multi-tasks.

\begin{table}[h]
\centering
\scriptsize
\begin{tabular}{lcccc}
\toprule
\textbf{Label} & \textbf{Precision} & \textbf{Recall} & \textbf{F1-score} & \textbf{Support} \\
\midrule
\multicolumn{5}{c}{\textbf{$\texttt{HypER}_{1-hop}$}} \\
\midrule
 Score 0 & 0.92   &   0.79   &   0.85    &   574 \\
 Score 1  & 0.13   &   0.57   &   0.22    &    42 \\
 Score 2 & 0.77   &   0.54   &   0.64     &  203 \\
\cmidrule{2-5}
 \textbf{Accuracy} & \multicolumn{3}{c}{\textbf{72.04\%}} & 819 \\
 \textbf{Macro Avg} &  0.61   &   0.64  &    0.57   &    819 \\
 \textbf{Weighted Avg} & 0.84  &    0.72   &   0.77   &    819 \\
\midrule
\multicolumn{5}{c}{\textbf{$\texttt{HypER}_{multi-hop-A}$}} \\
\midrule
 Invalid & 0.81    &  0.91   &   0.86   &    253 \\
 Valid   & 0.90   &   0.79   &   0.84   &    257 \\
\cmidrule{2-5}
 \textbf{Accuracy} & \multicolumn{3}{c}{\textbf{84.71\%}} & 510 \\
 \textbf{Macro Avg} & 0.85    &  0.85   &   0.85   &    510 \\
 \textbf{Weighted Avg} & 0.85   &   0.85   &   0.85   &    510 \\
\midrule
\multicolumn{5}{c}{\textbf{$\texttt{HypER}_{multi-hop-C}$}} \\
\midrule
 Invalid & 0.82  &    0.91  &    0.86    &   252 \\
 Valid   & 0.90   &   0.80  &    0.85   &    257 \\
\cmidrule{2-5}
 \textbf{Accuracy} & \multicolumn{3}{c}{\textbf{85.66\%}} & 509 \\
 \textbf{Macro Avg} & 0.86  &    0.86  &    0.86   &    509 \\
 \textbf{Weighted Avg} & 0.86   &   0.86   &   0.86   &    509 \\
\bottomrule
\end{tabular}
\caption{Detailed per-class classification performance for all classification tasks in \texttt{HypER}\_{Phi-3} evaluation. 
}
\label{tab:appendix_classification}
\end{table}

\section{Impact of Chain length}
\begin{figure}
    \centering
        \centering
        \includegraphics[width=\linewidth]{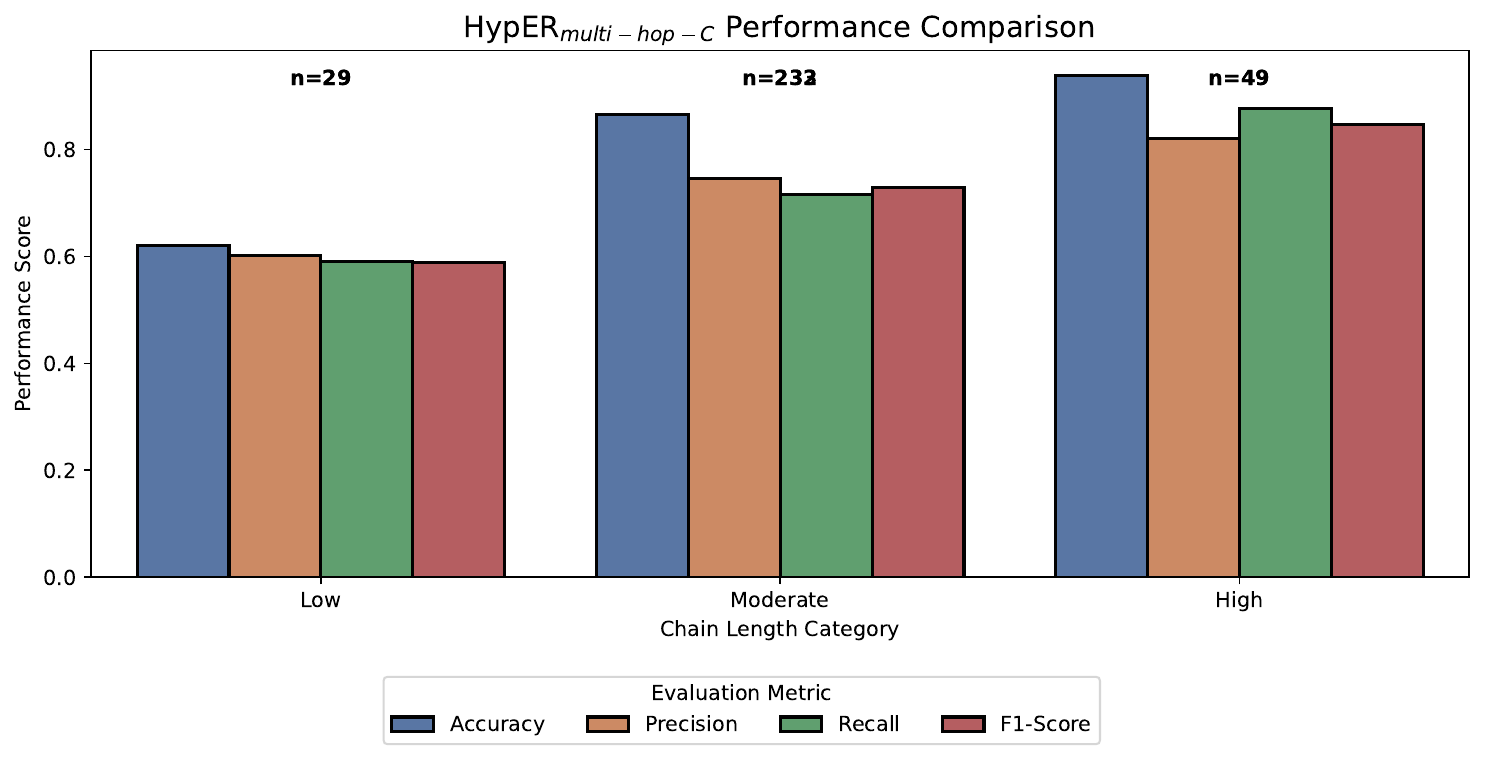}
    \caption{Classification performance (\texttt{HypER}_{Phi3}) on test data with categories of chain (Low, Moderate, and High) based on length.}
    \label{fig:performance_length}
\end{figure}

We also tested how chain length influenced \texttt{HypER}'s performance.
We expected that longer reasoning chains would capture more complex research progressions but might introduce irrelevant information. We analyze how \texttt{HypER} performs across small, moderate, and long chains in the reasoning chain validation task. We observe that \texttt{HypER} performance of chain validity classification achieves highest stability with moderate and longer chains (6+ papers), as shown in Figure~\ref{fig:performance_length}. For moderate-length chains (6-15 papers), the F1-score reaches 72.96\%, while shorter chains ($\leq$ 5 papers) perform worse at 58.94\% due to limited context. Longer chains ($\geq$ 16 papers) achieve the highest overall F1-score (84.64\%), indicating additional context strengthens reasoning validation rather than excessive noise. Despite expectations, \texttt{HypER} maintains strong recall on longer chains suggesting richer context helps in validation rather than hindering it.

\section{Human evaluation analysis}\label{appendix: expert_analysis}

\begin{figure*}[htbp!]
     \centering
     \includegraphics[width=\textwidth]{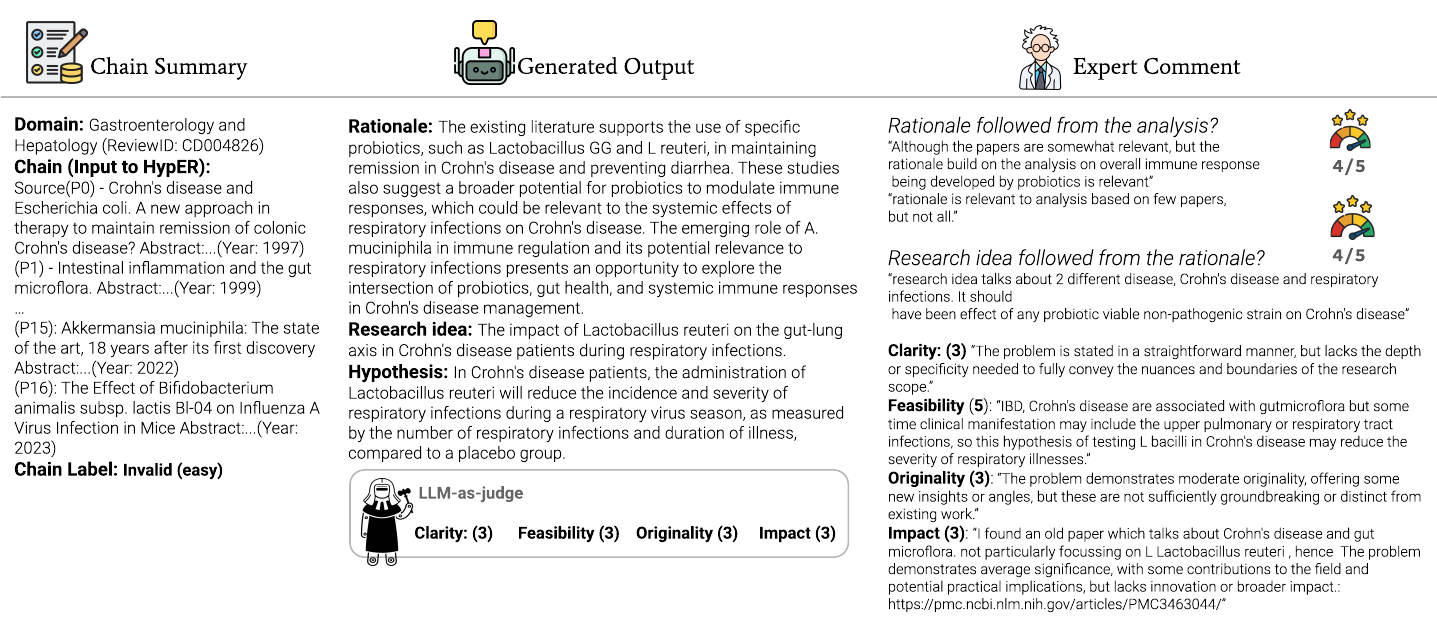}
     \caption{Example of a hypothesis generated by \texttt{HypER} from an invalid reasoning chain. The generated rationale, research idea, and hypothesis are evaluated by both LLM-as-judge and human experts. While the output demonstrates moderate clarity and feasibility, expert feedback highlights limitations in grounding, originality, and alignment with the cited literature. The input prompt used for generation is provided in Listing~\ref{appendix:prompt_hg}.}
     \label{fig:unsat_expert_example}
\end{figure*}

\subsection{Recruiting experts}
We recruited 10 medical and scientific experts through Upwork, a global freelancing platform that enabled us to identify professionals with relevant expertise in healthcare, clinical research, and scientific analysis. The evaluators included Doctors of Medicine (MDs) (4), Biomedical Scientists (2), Pharmaceutical Researchers (2), and Public Health Experts (2), whose expertise spanned clinical practice, biomedical research, pharmaceutical regulation, and scientific content evaluation. To verify their suitability, we had multiple conversations with experts and provided them with sample tasks (e.g., evaluation criteria along with an example-generated hypothesis and details as mentioned in section~\ref{sec:experiments} and Listing~\ref{app:study_instruction}). This process helped us confirm that they were well-equipped to perform the evaluation.  These diverse backgrounds ensured a useful and reliable evaluation of \texttt{HypER}’s generated hypotheses and explanations, and make our evaluation unusually thorough compared with studies that have used non-experts (even the paper authors themselves) to attempt to judge hypothesis quality.

\subsection{Expert evaluation analysis}
Our human evaluation of HypER's results provided several noteworthy insights that highlight both its strengths and areas for further improvement:

The expert analysis for the example output generated in 
Figure~\ref{fig:sat_expert_example}
highlights that the first two papers effectively build on the original study by demonstrating the impact of nurse-CHW interventions on diabetes management, while Papers 3 and 4 focus on unrelated topics. The rationale accurately reflects this model’s evaluation, and the proposed research idea logically explores the long-term effectiveness of these interventions. The expert provided a recent 2024 study (to which the model did not have access) that indicates that ongoing research trends align closely with the generated hypothesis, supporting its feasibility and relevance.

In the second example output generated in 
Figure~\ref{fig:unsat_expert_example}, the expert acknowledged that the rationale appropriately connected probiotics, immune response, and gut health but noted that it was only partially supported by the referenced papers in the chain. The research idea, exploring the impact of \textit{L. reuteri} on the gut-lung axis in Crohn's disease patients, was critiqued for combining unrelated conditions, suggesting that a more focused approach would be preferable. The hypothesis was considered feasible but was assessed as having moderate originality and average significance, with limited innovation. The expert also referenced an earlier study (\url{https://pmc.ncbi.nlm.nih.gov/articles/PMC3463044/}) that discussed gut microflora and Crohn's disease but did not specifically address the proposed focus on \textit{L. reuteri}.

\subsection{Expert vs. LLM Evaluation}\label{appendix:llm_human_correlation}

\begin{figure}[htbp!]
     \centering
     \includegraphics[width=\linewidth]{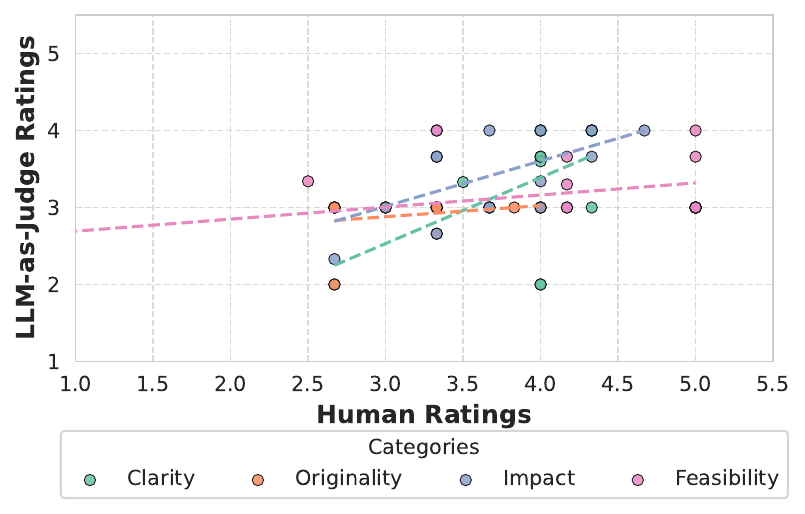}
     \caption{LLM vs. Human Ratings}
     \label{fig:human_vs_llm}
 \end{figure}

Table~\ref{tab:human_vs_llm} shows that expert and LLM-as-judge ratings broadly correlate across evaluation dimensions. The scoring protocol used by the experts and judge agent is given in Listing~\ref{scoring_protocol}. Human ratings are consistently higher, particularly in feasibility and clarity. To quantify this alignment, we computed Pearson and Spearman correlations across 15 examples (see Table~\ref{tab:human_llm_correlation} and also Figure~\ref{fig:human_vs_llm}). While clarity ($r=0.53$, $\rho=0.57$) and impact ($r=0.57$, $\rho=0.51$) show moderate correlation, originality and feasibility exhibit weaker agreement, with feasibility showing near-zero rank correlation ($\rho = -0.08$). These results suggest that LLMs may favor surface-level fluency over scientific plausibility, leading to occasional divergence from expert judgment, especially on harder or noisier examples.

\begin{table}[t]
\centering
\small
\begin{tabular}{lcc}
\toprule
\textbf{Metric} & \textbf{Expert Rating} & \textbf{LLM-as-judge Rating} \\
\midrule
Clarity         & $3.88 \pm 0.47$ & $3.28 \pm 0.76$ \\
Originality     & $3.21 \pm 0.50$ & $2.91 \pm 0.27$ \\
Feasibility     & $4.22 \pm 1.20$ & $3.20 \pm 0.50$ \\
Impact          & $3.69 \pm 0.54$ & $3.42 \pm 0.56$ \\
\bottomrule
\end{tabular}
\caption{Comparison of expert and LLM-as-judge evaluation scores ($\mu \pm \sigma$) for hypotheses generated by \texttt{HypER}\textsubscript{Phi-3}.}
\label{tab:human_vs_llm}
\end{table}

\begin{table}[t]
\centering
\small
\begin{tabular}{l@{\hskip 5pt}c@{\hskip 5pt}c}
\toprule
\textbf{Metric} & \textbf{Pearson Correlation} & \textbf{Spearman Correlation} \\
\midrule
Clarity      & 0.531 & 0.567 \\
Originality  & 0.270 & 0.183 \\
Impact       & 0.573 & 0.510 \\
Feasibility  & 0.376 & $-0.082$ \\
\bottomrule
\end{tabular}
\caption{Correlation between human and LLM-as-judge ratings across evaluation dimensions. While clarity and impact show moderate alignment, originality and feasibility demonstrate weaker or inconsistent agreement.}
\label{tab:human_llm_correlation}
\end{table}

\section{Supplementary Comparison}\label{appendix:auto_eval_LLM}
Due to space constraints, we provide an automated comparison between HypER and GPT-4o to evaluate the quality of generated hypotheses here. This comparison follows the same LLM-as-judge scoring protocol described in Listing~\ref{scoring_protocol}, adapted from~\cite{baek2024researchagent}.

\paragraph{Evaluation Protocol.} Both GPT-4o and HypER\_Phi3 were prompted with the same input reasoning chains from the test set, and their generated hypotheses were rated using LLM-as-judge framework (based on GPT-4). Each hypothesis was scored along the five axes defined in the scoring rubric: \textbf{Clarity}, \textbf{Relevance}, \textbf{Originality}, \textbf{Feasibility}, and \textbf{Significance}, using 5-point Likert-scale ratings. 

\subsection{Comparison with GPT-4o}\label{appendix:auto_eval_gpt4}
While this evaluation protocol may confer an advantage to GPT-4o, HypER (backboned by \texttt{Phi-3-mini-128k-instruct-3.8B}) still achieved comparable scores in key dimensions such as originality and significance.

\begin{table}[H]
\centering
\small
\begin{tabular}{lcc}
\toprule
\textbf{Dimension} & \textbf{GPT-4o} & \textbf{HypER\_Phi-3} \\
\midrule
Clarity        & 4.00 & 3.35 \\
Relevance      & 3.96 & 3.35 \\
Originality    & 3.00 & 3.01 \\
Feasibility    & 3.53 & 3.15 \\
Significance   & 3.84 & 3.37 \\
\bottomrule
\end{tabular}
\caption{Automated LLM-as-judge evaluation of hypotheses generated by GPT-4o and HypER.}
\label{tab:gpt4-compare}
\end{table}

From these results, GPT-4o outputs are slightly clearer and more relevant, while HypER (built on Phi-3-mini with only 3.8B parameters) performs comparably in originality and significance - suggesting that distilled models can approach SOTA. 

\subsection{Baseline Comparison: Abstract-only Phi-3}\label{appendix:auto_eval_source_base}
To further isolate the contribution of HypER's reasoning-driven generation, we compared it against the base Phi-3 model prompted only with the source abstract (i.e., without any intermediate reasoning chain). This setup reflects a naive baseline without explicit guidance. The same LLM-as-judge evaluation protocol was applied. We provide a dimension-wise analysis in Table~\ref{tab:phi3-abstract-compare}:

\begin{table}[H]
\centering
\small
\begin{tabular}{lcc}
\toprule
\textbf{Dimension} & \textbf{Phi-3 (Abstract-only)} & \textbf{HypER\_Phi-3} \\
\midrule
Clarity        &  3.80  & 3.35 \\
Relevance      & 3.72 & 3.35 \\
Originality    & 2.95  & 3.01 \\
Feasibility    & 3.28 & 3.15 \\
Significance   & 3.69  & 3.37 \\
\bottomrule
\end{tabular}
\caption{LLM-as-judge evaluation comparing naive Phi-3 (abstract-only) and HypER\_Phi-3 (reasoning-guided).}
\label{tab:phi3-abstract-compare}
\end{table}

While abstract-only baseline LLM can generate creative, open-ended hypotheses, they often lack reasoning over literature and identification of evidence gaps. The abstract-only baseline reflects this, producing fluent but generic outputs. In contrast, HypER uses validated reasoning chains to generate more original, evidence-backed hypotheses aligned with our goal of structured, literature-grounded generation.

\subsection{Focus of Evaluation}
While we report scores for all five dimensions for completeness, we emphasize that \textbf{our primary objective is not to optimize for originality}. HypER is fine-tuned to identify coherent, evidence-backed reasoning chains and generate hypotheses grounded in scientific dependencies, rather than unconstrained novelty. Thus, its strength lies not in stylistic surface quality but in its ability to reason over noisy literature graphs and produce outputs that are more meaningful in scientific contexts.

Importantly, what we obtain from HypER is not just a hypothesis, but a full explanation that \textit{analyzes the reasoning chain, identifies knowledge gaps, and formulates a research idea as a specific, evaluable hypothesis}. Therefore, optimizing solely for surface-level dimensions such as clarity or fluency-where large language models like GPT-4o may have an inherent advantage-would defeat the core purpose of HypER. Our design objective is not stylistic polish, but faithful alignment with scientific reasoning and tractable research generation.

A more thorough evaluation using diverse LLM-as-judge setups or human expert feedback could be a valuable future direction, but is considered out of scope for this work.

\section{Study Instructions and Prompts}
Detailed study instruction we have used for the expert evaluation is provided in Listing~\ref{app:study_instruction}. The scoring protocol used by the experts and judge agent is given in Listing~\ref{scoring_protocol}.

The prompts we used for llama relevancy scoring is detailed in Listing~\ref{appendix:prompt_llama_score}. The hypotheses generation prompt is illustrated in Listing~\ref{appendix:prompt_hg}, and the prompt used for the judge agent is shown in Listing~\ref{appendix:prompt_judge_agent}.

\begin{table*}
\begin{tcolorbox}[
    colback=purple!10,             
    colframe=purple!80,            
    coltitle=white,                
    colbacktitle=purple!80,        
    fonttitle=\bfseries\large,     
    title=Example of a valid reasoning chain,  
    rounded corners,               
    boxrule=0.75mm                 
]
\small  

\textbf{Title:} Evidence suggesting that a chronic disease self-management program can improve health status while reducing hospitalization \newline
\textbf{Abstract:} This study evaluated the effectiveness (changes in health behaviors, health status, and health service utilization) of a self-management program for chronic disease ... \newline
\textbf{Year:} 1999 \newline
\textbf{Citation Count:} 2315 \newline
\textbf{Relevance:} -- (Source Paper) \newline

\textbf{Title:} Effectiveness of a community health worker intervention among African American and Latino adults with type 2 diabetes \newline
\textbf{Abstract:} We tested the effectiveness of a culturally tailored, behavioral theory-based community health worker intervention for improving glycemic... \newline
\textbf{Year:} 2011 \newline
\textbf{Citation Count:} 332 \newline
\textbf{Relevance:} 2 \newline
\textbf{explanation:} This paper is partially dependent on the findings of the source paper, as it investigates the effectiveness of a community health worker intervention for improving glycemic control, which is a related topic to the source paper's focus on self-management programs for chronic disease patients. \newline

\textbf{Title:} Nurse–Community Health Worker Team Improves Diabetes Care in American Samoa \newline
\textbf{Abstract:} To evaluate the effectiveness of a culturally adapted, primary care-based nurse community health worker (CHW) team intervention to support diabetes self-management on diabetes control ... \newline
\textbf{Year:} 2013 \newline
\textbf{Citation Count:} 100 \newline
\textbf{Relevance:} 2 \newline
\textbf{explanation:} The key hypothesis in this paper is at least partially dependent on the findings of the source paper, as it evaluates the effectiveness of a nurse-community health worker team in improving diabetes care, building on the source paper's results regarding community health worker interventions for diabetes management. \newline

\textbf{Title:} Impact of a diabetes control and management intervention on health care utilization in American Samoa \newline
\textbf{Abstract:}  To examine the impact of a successful 12-month behavioral intervention to improve diabetes control on health care utilization in American Samoa.... \newline
\textbf{Year:} 2014 \newline
\textbf{Citation Count:} 17 \newline
\textbf{Relevance:} 2 \newline
\textbf{explanation:} This paper examines the impact of a successful 12-month behavioral intervention to improve diabetes control on health care utilization in American Samoa, and builds upon the source paper's findings on the effectiveness of a culturally adapted nurse-community health worker team intervention in improving diabetes control. \newline

\textbf{Title:} Diabetes training for community health workers on an American Indian reservation \newline
\textbf{Abstract:} A quality improvement program aimed at enhancing the knowledge and skills of community health workers in managing diabetes through formal training... \newline
\textbf{Year:} 2018 \newline
\textbf{Citation Count:} 16 \newline
\textbf{Relevance:} 2 \newline
\textbf{explanation:} This paper is closely related to the source paper, as it focuses on training community health workers to improve diabetes management, which aligns with the source paper's intervention. Moreover, the paper's emphasis on community health worker training can be seen as a sub-hypothesis of the source paper's findings on the effectiveness of a community health worker diabetes intervention. Therefore, this paper is at least partially dependent on the findings of the source paper. \newline

\textbf{Title:} What Is the Impact of a Context-Specific Training Program for Home-Based Carers? An Evaluation Study \newline
\textbf{Abstract:} This study evaluated a training program for home-based carers in South Africa, aiming to improve diabetes care knowledge and skills... \newline
\textbf{Year:} 2020 \newline
\textbf{Citation Count:} 2 \newline
\textbf{Relevance:} 1 \newline
\textbf{explanation:} This paper is inspired by the hypothesis of the source paper, as it also investigates the impact of training on the knowledge and skills of community health workers, specifically home-based carers, in managing diabetes. \newline

\end{tcolorbox}
    \caption{Example reasoning chain with the relevancy score computed and the corresponding explanation}
    \label{tab:dependency_chain}
\end{table*}

\begin{table*}
\begin{tcolorbox}[
    colback=purple!10,             
    colframe=purple!80,            
    coltitle=white,                
    colbacktitle=purple!80,        
    fonttitle=\bfseries\large,     
    title=Example of an invalid reasoning chain,  
    rounded corners,               
    boxrule=0.75mm                 
]
\small  

\textbf{Title:} Lactobacillus paracasei feeding improves immune control of influenza infection in mice \newline
\textbf{Abstract:} Respiratory tract infections such as flu cause severe morbidity and mortality and are among the leading causes of death in children and adults worldwide. Commensal microbiota ... \newline
\textbf{Year:} 2017 \newline
\textbf{Citation Count:} 80 \newline
\textbf{Relevance:} -- (Source Paper) \newline

\textbf{Title:} Akkermansia muciniphila Improves Host Defense Against Influenza Virus Infectio \newline
\textbf{Abstract:} Influenza virus infection can alter the composition of the gut microbiota, while its pathogenicity can, in turn,... \newline
\textbf{Year:} 2021 \newline
\textbf{Citation Count:} 44 \newline
\textbf{Relevance:} 1 \newline
\textbf{explanation:} This paper explores the effects of Akkermansia muciniphila on influenza virus infection, which is related to the source paper's topic of using probiotics (Lactobacillus delbrueckii ssp. bulgaricus OLL1073R-1) to enhance humoral immune responses against influenza A virus. However, the connection is not direct, and the paper does not build upon the source paper's findings \newline

\textbf{Title:} Akkermansia muciniphila: The state of the art, 18 years after its first discovery \newline
\textbf{Abstract:} Akkermansia muciniphila (A. muciniphila) is an anaerobic, Gram negative and mucin-degrading bacterium of the phylum... \newline
\textbf{Year:} 2022 \newline
\textbf{Citation Count:} 9 \newline
\textbf{Relevance:} 0 \newline
\textbf{explanation:} This paper is a review of the current state of knowledge on Akkermansia muciniphila, which is the main subject of the source paper. However, it does not build upon the findings of the source paper or propose a new hypothesis inspired by it. \newline

\end{tcolorbox}
    \caption{Example reasoning chain (invalid) with the relevancy score computed and the corresponding explanation.}
    \label{tab:dependency_chain2}
\end{table*}

\begin{table*}
\begin{tcolorbox}[
    colback=blue!10,             
    colframe=blue!80,            
    coltitle=white,                
    colbacktitle=blue!80,        
    fonttitle=\bfseries\large,     
    title=Example of generated results,  
    rounded corners,               
    boxrule=0.75mm                 
]
\small  

\textbf{"generated_research_idea"} : \{
\textbf{"Analysis":} 
\textbf{'0':} 'The paper by Pharmacological interventions for peripheral artery disease (2007) discusses the need for well-designed protocols to evaluate the effectiveness of various treatments for PAD, including exercise training, smoking cessation, and antiplatelet therapy. This paper sets the stage for evaluating the effectiveness of buflomedil in diabetic subjects with intermittent claudication by providing a framework for future studies.',\\
\textbf{'1':} 'The Importance of Conservative Measures in Peripheral Arterial Disease: An Update (2008) does not directly build upon the previous papers but emphasizes the importance of conservative measures in managing PAD. It highlights the need for further research on the effectiveness of pharmacological interventions, including buflomedil, in improving walking distance and reducing cardiovascular risk.',\\
\textbf{'2':} 'Ankle\u2014Brachial Index: A Marker of Both Peripheral Arterial Disease and Systemic Atherosclerosis As Well As a Predictor of Vascular Events (2010) does not directly build upon the previous papers but underscores the significance of the ankle-brachial index as a diagnostic tool for PAD.',\\
\textbf{'3':} 'Contemporary Management of Infrapopliteal Peripheral Arterial Disease (2011) does not directly build upon the previous papers but reviews the historical challenges and limitations of surgical and endovascular interventions for infrapopliteal PAD.',\\
'4': 'Drug eluting stents in the treatment of below the knee arterial occlusive disease (2013) is not relevant to the research topic and should be ignored.',\\
\textbf{'5':} 'Initial Experience With the Absorb Bioresorbable Vascular Scaffold in Infrapopliteal Artery Disease (2015) and Early and Midterm Experience With the Absorb Everolimus-Eluting Bioresorbable Vascular Scaffold in Asian Patients With Chronic Limb-Threatening Ischemia (2020) discuss the use of a bioresorbable vascular scaffold for treating infrapopliteal artery disease. These papers build upon the previous research by demonstrating the feasibility and safety of using this device in patients with critical limb ischemia.',\\
\textbf{'6':} 'Lessons Learned Regarding Handling and Deployment of the Absorb Bioresorbable Vascular Scaffold in Infrapopliteal Arteries (2017) provides technical details and considerations for the successful use of the Absorb BVS in infrapopliteal artery disease. It builds upon the previous papers by offering insights into the practical aspects of deploying this device in a clinical setting.',\\
\textbf{'7':} 'Three-Year Results of the Absorb Everolimus-Eluting Bioresorbable Vascular Scaffold in Infrapopliteal Arteries (2018) reports the midterm performance of the Absorb BVS in patients with infrapopliteal artery disease. It builds upon the previous papers by providing long-term follow-up data on safety and efficacy.',\\
\textbf{'8':} 'Early and Midterm Experience With the Absorb Everolimus-Eluting Bioresorbable Vascular Scaffold in Asian Patients With Chronic Limb-Threatening Ischemia (2020) provides additional data on the performance of the Absorb BVS in a diverse patient population. It builds upon the previous papers by contributing to the understanding of the device's effectiveness in different clinical scenarios.',\\
\textbf{'9':} 'Drug-Eluting Resorbable Scaffold versus Angioplasty for Infrapopliteal Artery Disease (2023) compares the outcomes of using a drug-eluting resorbable scaffold versus angioplasty in patients with infrapopliteal artery disease. It builds upon the previous papers by providing a direct comparison of the two treatment modalities in a randomized controlled trial.'",
\\
\textbf{"Rationale":} "The progression of research on the treatment of infrapopliteal artery disease has evolved from evaluating the clinical and hemorheological effects of buflomedil (source paper) to assessing the feasibility and safety of using bioresorbable vascular scaffolds (BVS) for this condition. The focus has shifted towards comparing the long-term outcomes and effectiveness of different treatment options, including drug-eluting BVS, in patients with critical limb ischemia. This progression highlights the need for high-quality evidence to guide treatment decisions in this challenging patient population. To advance the field, a research idea that builds upon these studies while addressing their limitations and incorporates novel aspects is needed."
\\
\textbf{"Research idea":} "Feasibility and safety assessment of drug-eluting BVS in patients with critical limb ischemia due to infrapopliteal artery disease. This research problem aims to provide high-quality evidence on the effectiveness of drug-eluting BVS compared to angioplasty in improving clinical outcomes and limb salvage rates in patients with critical limb ischemia. The study will also assess the feasibility of using drug-eluting BVS in complex, calcified lesions."
\\
\textbf{ "Hypothesis":} "In patients with critical limb ischemia due to infrapopliteal artery disease, drug-eluting BVS will result in higher primary patency rates, lower rates of major adverse limb events, and improved limb salvage rates compared to angioplasty at 1 year. The feasibility of using drug-eluting BVS in complex, calcified lesions will be assessed based on the rate of successful implantation and technical success."
\}
\end{tcolorbox}
    \caption{Example of generated analysis, Rationale, Research idea and Hypothesis for a given source paper and reasoning chain.}
    \label{tab:generated_hypothesis}
\end{table*}

\clearpage
\onecolumn

\label{appendix:prompt_llama_score}
\begin{tcolorbox}[
    colback=gray!10, 
    colframe=gray!80!black, 
    coltitle=white,
    enhanced,
    breakable,
    listing only,
    sharp corners,
    boxrule=0.7mm,
    title=Relevancy Scoring Prompt,
    width=\textwidth,
]
\begin{minted}{text}

|\textbf{system}| = f"""You are a helpful assistant designed to evaluate scientific literature.""" 
|\textbf{user}| = f""" Hypotheses are frequently the starting point when undertaking the empirical portion of the scientific process. They state something that the scientific process will attempt to evaluate, corroborate, verify, or falsify. Their purpose is to guide the types of data we collect, analyses we conduct, and inferences we would like to make. You are a scientist. Your job is to construct a novel and impactful hypothesis by navigating the literature.

    We have retrieved a knowledge graph of literature for you. You are given a source paper and a list of papers that followed from the source paper.
    You are evaluating the relevance of the following papers to the source paper. Starting from the source paper, you will analyze the following papers in this way. For every paper in the list, you output 0, 1, 2:
    0: This paper has no connection with the source paper or this paper is a review paper (e.g., Cochrane reviews, systematic reviews). Review papers often include terms like "Review" or "Meta-Analysis," summarize existing literature, and lack novel hypotheses or findings.
    1: The key hypothesis in this paper is inspired by the hypothesis or the finding from the source paper
    2: The key hypothesis in this paper is at least partially dependent on the findings of the source paper. In other words the source papers contain some sub-hypotheses for the current hypothesis.

    Explain your answer.

    If there are 5 papers, your answer should contain an enumerated list of length 5. 
    
    Finally, identify the top-3 relevant papers from the list based on the highest relevance score (2 > 1 > 0). If there are fewer than 3 most relevant papers (with scores 1 or 2), include only the available ones. If no relevant papers are found, leave the "top3_relevant_papers" section empty.

    Few-shot examples:
    {few_shot_prompt}

    Source Paper:
    Title: {source_title}
    Abstract: {source_abstract}

    Papers from the Year {year}:
    {paper_list}

    Output a JSON object in the following format:
    ```json
    {{
        "paper_list": {{
            "1.Title of the First Paper": {{
                "explanation": "Explanation of the connection to the source paper.",
                "relevance": 0, 1, or 2
            }},
            "2.Title of the Second Paper": {{
                "explanation": "Explanation of the connection to the source paper.",
                "relevance": 0, 1, or 2
            }},
            ...
        }},
        "top3_relevant_papers": {{
            "1.title of the first relevant paper": {{
                "explanation": "Explanation of the connection to the source paper.",
                "relevance": 1, or 2
            }},
            "2.Title of the second relevant paper": {{
                "explanation": "Explanation of the connection to the source paper.",
                "relevance": 1, or 2
            }},
            "3.Title of the third relevant paper": {{
                "explanation": "Explanation of the connection to the source paper.",
                "relevance": 1, or 2
            }}
    }}
    ```
"""

\end{minted}
\end{tcolorbox}
\captionof{listing}{Prompt for Llama relevancy scoring}

\label{app:study_instruction}
\begin{tcolorbox}[
    colback=gray!10, 
    colframe=gray!80!black, 
    coltitle=white,
    enhanced,
    breakable,
    listing only,
    sharp corners,
    boxrule=0.7mm,
    title=Evaluation instructions,
    width=\textwidth,
]
\begin{minted}{text}

|\textbf{Instructions}|

You are a reviewer whose primary goal is to assess the quality and validity of scientific research problems and hypothesis generated by AI. The AI was given a source paper and asked to generate novel and impactful research pertaining to the topic of the source paper. The AI was also given access to the literature in the form of a chain of papers arranged in temporal order. The chain could be noisy, so not all papers were directly relevant to the source paper. The AI was expected to consider the relevant papers and perform a short literature review which we call rationale and then come up with research idea based on the identified research gap. It was asked to generate a hypothesis that would be clear, novel, feasible or testable and impactful (if possible). Your goal is to understand if the AI did each of its task well. 

You will evaluate the following:

1. The first thing AI produced was an analysis of how each paper in the chain is relevant to the source paper. To help you judge better, we are putting the title and the abstract of each of the papers and the corresponding AI generated analysis in two adjacent columns. You will tell us if the judgment is correct or not and provide your comments.

2. Next AI generated a rationale which is supposed to be grounded in the above analysis. Your task is to judge whether the rationale is coherent i.e. whether it follows naturally from the analysis or whether the model is using significant amount of external knowledge to generate the rationale. External knowledge is something not contained in the provided papers. You will be given a few questions related to the quality of the rationale.

3. The rationale serves a motivation for a new research idea. The AI was asked to generate a research idea motivated by the rationale. Your task is to evaluate whether the research idea follows from the rationale.

4. Your last task is to evaluate the quality of the hypothesis. The purpose of the hypothesis is to translate the research idea into a concrete testable declarative statement. You will evaluate whether hypothesis is clear, testable, follows from the research idea and novel.

\end{minted}
\end{tcolorbox}
\captionof{listing}{Expert evaluation instructions}

\label{scoring_protocol}
\begin{tcolorbox}[
    colback=gray!10, 
    colframe=gray!80!black, 
    coltitle=white,
    enhanced,
    breakable,
    listing only,
    sharp corners,
    boxrule=0.7mm,
    title=Scoring protocol for Judge Agent~\citep{Baek2024ResearchAgentIR},
    width=\textwidth,
]
\begin{minted}{text}

{
    |\textbf{"clarity":}| "1. The problem is presented in a highly ambiguous manner, lacking clear definition and leaving significant room for interpretation or confusion.
    2. The problem is somewhat defined but suffers from vague terms and insufficient detail, making it challenging to grasp the full scope or objective.
    3. The problem is stated in a straightforward manner, but lacks the depth or specificity needed to fully convey the nuances and boundaries of the research scope.
    4. The problem is clearly articulated with precise terminology and sufficient detail, providing a solid understanding of the scope and objectives with minimal ambiguity.5. The problem is exceptionally clear, concise, and specific, with every term and aspect well-defined, leaving no room for misinterpretation and fully encapsulating the research scope and aims.",
    
    |\textbf{"relevance":}| "1. The problem shows almost no relevance to the current field, failing to connect with the established context or build upon existing work. 
    2. The problem has minimal relevance, with only superficial connections to the field and a lack of meaningful integration with prior studies. 
    3. The problem is somewhat relevant, making a moderate attempt to align with the field but lacking significant innovation or depth. 
    4. The problem is relevant and well-connected to the field, demonstrating a good understanding of existing work and offering promising contributions. 
    5. The problem is highly relevant, deeply integrated with the current context, and represents a significant advancement in the field.",
    
    |\textbf{"originality":}| "1. The problem exhibits no discernible originality, closely mirroring existing studies without introducing any novel perspectives or challenges. 
    2. The problem shows minimal originality, with slight variations from known studies, lacking significant new insights or innovative approaches. 
    3. The problem demonstrates moderate originality, offering some new insights or angles, but these are not sufficiently groundbreaking or distinct from existing work. 
    4. The problem is notably original, presenting a unique challenge or perspective that is well-differentiated from existing studies, contributing valuable new understanding to the field.
    5. The problem is highly original, introducing a pioneering challenge or perspective that has not been explored before, setting a new direction for future research.",

    |\textbf{"feasibility":}| "1. The problem is fundamentally infeasible due to insurmountable resource constraints, lack of foundational research, or critical methodological flaws. 
    2. The problem faces significant feasibility challenges related to resource availability, existing knowledge gaps, or technical limitations, making progress unlikely. 
    3. The problem is feasible to some extent but faces notable obstacles in resources, existing research support, or technical implementation, which could hinder significant advancements. 
    4. The problem is mostly feasible with manageable challenges in resources, supported by adequate existing research, and has a clear, achievable methodology, though minor issues may persist. 
    5. The problem is highly feasible with minimal barriers, well-supported by existing research, ample resources, and a robust, clear methodology, promising significant advancements.",

   |\textbf{"significance":}| "1. The problem shows minimal to no significance, lacking relevance or potential impact in advancing the field or contributing to practical applications. 
   2. The problem has limited significance, with a narrow scope of impact and minor contributions to the field, offering little to no practical implications. 
   3. The problem demonstrates average significance, with some contributions to the field and potential practical implications, but lacks innovation or broader impact. 
   4. The problem is significant, offering notable contributions to the field and valuable practical implications, with evidence of potential for broader impact and advancement.
   5. The problem presents exceptional significance, with groundbreaking contributions to the field, broad and transformative potential impacts, and substantial practical applications across diverse domains."
}

\end{minted}
\end{tcolorbox}
\captionof{listing}{Scoring Protocol borrowed from~\citep{Baek2024ResearchAgentIR}}

\label{appendix:prompt_hg}
\begin{tcolorbox}[
    colback=gray!10, 
    colframe=gray!80!black, 
    coltitle=white,
    enhanced,
    breakable,
    listing only,
    sharp corners,
    boxrule=0.7mm,
    title=Prompt for hypothesis generation,
    width=\textwidth,
]
\begin{minted}{text}

system_message = """You are an AI assistant whose primary goal is to identify promising, new, and key scientific problems based on existing scientific literature, in order to aid researchers in discovering novel and significant research opportunities that can advance the field."""

user_message = f"""You are going to generate a research problem that should be original, clear, feasible, relevant, and significant to its field. This will be based on the title and abstract of the source paper, those of {len(citing_paper_list)} related papers in the existing literature.
    Understanding of the target paper, and the related papers is essential:
    - The source paper is the primary research study you aim to enhance or build upon through future research, serving as the central source and focus for identifying and developing the specific research problem.
    - The related papers are arranged in temporal order of citation, such that paper 0 cites the source paper, 2 cites paper 1 and paper 3 cites paper 2 and so on. The relevant papers provide additional context and insights that are essential for understanding and expanding upon the target paper. However, all the papers in the list may not be relevant to the primary research you are focusing on. Identify the most relevant papers from the list in your analysis and only use those for research idea generation.
    Your approach should be systematic:
    - Start by thoroughly reading the title and abstract of the source paper to understand its core focus.
    - Next, proceed to read the titles and abstracts of the related papers in the order in which they appear in the list. 
    Identify the papers that form a logical reasoning chain starting from the source paper.
    - Use only these papers to gain a broader perspective about the progression of the primary research topic over time. 

    ###Example Task & Expected Output:
    ###Example Input:
    Source paper title: {one_shot['source paper']['title']}
    Source paper abstract: {one_shot['source paper']['abstract']}
    Source paper year of publication: {one_shot['source paper']['year']}
    Related papers: {one_shot['related papers']}

    ### Example Output (Valid JSON Format):
    ```json
    {{
    "Analysis": {one_shot['output']['<analysis>']},
    "Rationale": "{one_shot['output']['<motivation>']}",
    "Research idea": "{one_shot['output']['<research idea>']}",
    "Hypothesis": "{one_shot['output']['<hypothesis>']}"
    }}
    ```
    ###
    ### **Important:  **Do not copy from the example above.** Instead, based on the provided source and related papers to generate a research problem that should be original, clear, feasible, relevant, and significant to its field. 

    I am going to provide the source paper and related papers as an enumerated list of Title, Abstract and Year of publication triple, as follows:
    Source paper title: {source_paper['title']}
    Source paper abstract: {source_paper['abstract']}
    Source paper year of publication: {source_paper['year']}
    Related papers: {citing_paper_list}
    
    With the provided source paper, and the related papers, your objective now is to formulate a research problem that not only builds upon these existing studies but also strives to be original, clear, feasible, relevant, and significant. Before crafting the research problem, revisit the title and abstract of the source paper, to ensure it remains the focal point of your research problem identification process. Your research problem will be scored for clarity. It should contain a short description of the general research idea and it's impact followed by more details on all the variables and how they will be measured. 
    If possible include PICO elements which stands for Population, Intervention, Control and Outcome. 
    State clearly how the outcome could potentially be measured.

    Now convert this idea into a concrete testable hypothesis. Remember hypothesis is a declarative statement expressing a 
    relationship between two variables like independent or dependent variables or left group and right group in a given context.
    Your hypothesis should contain the key variable or variables from your research idea and how they will be measured.
    Your hypothesis will be scored on clarity and novelty.
    
    Source paper title: {source_paper['title']}
    Source paper abstract: {source_paper['abstract']}

    Then, following your review of the above content and example, please proceed to analyze the progression of the research topic. For analysis, Output a dictionary with each paper in the Related Papers as a key. For each key (paper) analyze how this paper builds upon the previous papers in the list. For example, how Paper 0 builds upon source paper and Paper 1 builds upon the concepts in Paper 0 and so on. Elaborate on specific advancements made, including the explanation behind their effectiveness in addressing previous challenges. Apply this analytical approach to each valid paper in the sequence, adding the analysis as the value for each key in a few sentences. Ignore papers that do not build upon the previous papers and diverge from the original source paper's topic significantly.
    
    Now output this analysis, the research problem and hypothesis with the rationale. Your output should be a valid JSON with the following fields.

    
    Output a JSON object in the following format
    ```json
    {{
    "Analysis": {{Output a dictionary with each paper in the Related Papers as a key. For each key (paper) analyze how this paper builds upon the previous papers in the list.}},
    "Rationale": "Summarize the above analysis and explain how you would come up with a research idea that will advance the field of work while addressing the limitations of previous work and building upon the existing work.",
    "Research idea": "Delineate an elaborate research problem here including the key variables.",
    "Hypothesis": "Provide a concrete testable hypothesis that follows from the above research problem here"
    }}
    ```
    
    This JSON will be automatically parsed, so ensure the format is precise. DO NOT leave any field empty. If you cannot generate a specific part, provide a best guess.
    """
\end{minted}
\end{tcolorbox}
\captionof{listing}{Prompt for hypothesis generation adapted from~\citep{baek2024researchagent}}

\label{appendix:prompt_judge_agent}
\begin{tcolorbox}[
    colback=gray!10, 
    colframe=gray!80!black, 
    coltitle=white,
    enhanced,
    breakable,
    listing only,
    sharp corners,
    boxrule=0.7mm,
    title= Prompts for Judge Agent,
    width=\textwidth,
]
\begin{minted}{text}

system_message = """You are an AI assistant whose primary goal is to assess the quality and validity of scientific problems across diverse dimensions, in order to aid researchers in refining their problems based on your evaluations and feedback, thereby enhancing the impact and reach of their work. Your response must be in JSON format"""
user_message = f"""You are going to evaluate a research problem for its {metric}, focusing on how well it is defined in a clear, precise, and understandable manner. As part of your evaluation, you can refer to the existing studies that may be related to the problem, which will help in understanding the context of the problem for a more comprehensive assessment.
- The existing studies refer to the target paper that has been pivotal in identifying the problem, as well as the related papers that have been additionally referenced in the discovery phase of the
problem.
The existing studies (target paper & related papers) are as follows:
Target paper title: {source_paper['title']}
Target paper abstract: {source_paper['abstract']}
Related papers: {related_papers}

Now, proceed with your {metric} evaluation approach that should be systematic:
- Start by thoroughly reading the research problem and its rationale, keeping in mind the context provided by the existing studies mentioned above.
- Next, generate a review and feedback that should be constructive, helpful, and concise, focusing on the {metric} of the problem.
- Finally, provide a score for the Hypothesis on a 5-point Likert scale, with 1 being the lowest. Be a harse critic. Please ensuring a discerning and critical evaluation and avoid uniformly high ratings (4-5) unless fully justified.

Following are the judging criteria for each rating number:
{scoring_protocol[metric]}

     ### Example Input & Expected Output}
     ### Example Input:}
     ###
     Example Input for Rating 1:
    Research problem: {fewshots['Rating 1']["Research problem"]}
    Rationale: {fewshots['Rating 1']['Rationale']}
    Hypothesis: {fewshots['Rating 1']['Hypothesis']}
    #### Example Output (Valid JSON Format):
    ```json
    {{
    "Review": {fewshots['Rating 1']['Review']},
    "Feedback": {fewshots['Rating 1']['Feedback']}"
    "Rating (1-5) for Hypothesis": 1
    }}
    ```
    ###
    Example Input for Rating 3:
    Research problem: {fewshots['Rating 3']["Research problem"]}
    Rationale: {fewshots['Rating 3']['Rationale']}
    Hypothesis: {fewshots['Rating 3']['Hypothesis']}
    #### Example Output (Valid JSON Format):
    ```json
    {{
    "Review": {fewshots['Rating 3']['Review']},
    "Feedback": {fewshots['Rating 3']['Feedback']}
    "Rating (1-5) for Hypothesis": 3
    }}
    ```
    ###
    ###
    Example Input for Rating 5:
    Research problem: {fewshots['Rating 5']["Research problem"]}
    Rationale: {fewshots['Rating 5']['Rationale']}
    Hypothesis: {fewshots['Rating 5']['Hypothesis']}
    
    ### Example Output (Valid JSON Format):
    ```json
    {{
    "Review": "{fewshots['Rating 5']['Review']}",
    "Feedback": "{fewshots['Rating 5']['Feedback']}"
    "Rating (1-5) for Hypothesis": 5
    }}
    ```
    
    ###

I am going to provide the research problem with its rationale, as follows:
Research problem: {research_idea['Research idea']}
Rationale: {research_idea['Rationale']}
Hypothesis: {research_idea['Hypothesis']}
After your evaluation of the above content, please provide your review, feedback, and rating. 
Your output should be structured as follows:
RESPONSE:
```json
<JSON>
```
In <JSON>, respond in JSON format with ONLY the following field:
- "Review": Your review of the research problem.
- "Feedback": Your constructive feedback for improvement.
- "Rating (1-5) for Hypothesis": only output a rating number here.
This JSON will be automatically parsed, so ensure the format is precise.
"""

\end{minted}
\end{tcolorbox}
\captionof{listing}{Prompts for Judge Agent}

\twocolumn

\end{document}